\documentclass[11pt]{article}
\ifdefined\pdfminorversion
\fi
\usepackage[preprint]{acl}
\usepackage{times}
\usepackage{latexsym}
\usepackage[T1]{fontenc}
\usepackage[utf8]{inputenc}
\usepackage{microtype}
\usepackage{inconsolata}
\usepackage{listings}
\lstnewenvironment{promptblock}{\lstset{
  basicstyle=\small\ttfamily,
  columns=fullflexible,
  keepspaces=true,
  showstringspaces=false,
  breaklines=true,
  breakatwhitespace=true,
  breakindent=0pt,
  frame=single,
  framerule=0.3pt,
  framesep=6pt,
  xleftmargin=7pt,
  xrightmargin=7pt,
  aboveskip=0.6\baselineskip,
  belowskip=0.8\baselineskip
}}{}
\usepackage{graphicx}
\usepackage{amsmath}
\usepackage{algorithm}
\usepackage{algpseudocode}
\usepackage{booktabs}
\usepackage{colortbl}
\usepackage{placeins}

\title{\texorpdfstring{\raisebox{-0.25em}{\includegraphics[height=1.55em]{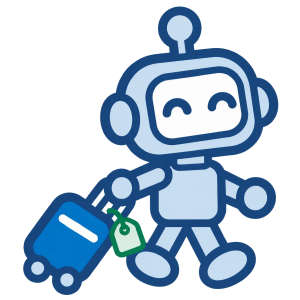}}\hspace{0.35em}}{}CarryOn: Learning What to Keep in Context\\
for Long-Horizon LLM Agents}
\author{
  Lihuan Li\textsuperscript{1}\quad
  Ruiyi Yang\textsuperscript{1}\thanks{\protect\hypertarget{Hfootnote.1}{}Ruiyi Yang and Jiru Wang contributed equally as co-second authors.}\quad
  Jiru Wang\textsuperscript{1}\hyperlink{Hfootnote.1}{\textsuperscript{\normalfont *}}\quad
  Yifan Duan\textsuperscript{1}\\
  \bfseries Xiachong Lin\textsuperscript{1}\quad
  Hao Xue\textsuperscript{2}\quad
  Yang Song\textsuperscript{1}\quad
  Flora Salim\textsuperscript{1}\\
  \normalfont\textsuperscript{1}University of New South Wales, Sydney, Australia\\
  \normalfont\textsuperscript{2}The Hong Kong University of Science and Technology (Guangzhou), Guangzhou, China\\
  {\small\ttfamily\{lihuan.li,ruiyi.yang,jiru.wang,yifan.duan,xiachong.lin\}@unsw.edu.au}\\
  {\small\ttfamily\{yang.song1,flora.salim\}@unsw.edu.au}\\
  {\small\ttfamily haoxue@hkust-gz.edu.cn}
}

\begin{document}
\ifdefined\XeTeXversion

\fi
\maketitle

\begin{abstract}
Long-horizon LLM agents can coordinate tools and applications to complete
complex, multi-step tasks. As their interaction histories grow, effective
context management becomes essential for reliable and efficient execution.
Existing methods reduce context through summarization, pruning, or history
selection. However, shorter contexts do not necessarily support better
decisions: summaries can omit exact details needed for later actions, while
retaining entire interactions can carry substantial irrelevant information.
We propose CarryOn, a lightweight context manager that pairs compact
execution state with fine-grained, source-linked evidence. A learned
selector and adaptive retention policy keep relevant evidence available
without additional LLM calls for context management.
We introduce two complementary evaluation metrics: $S_\alpha$ prioritizes
task success while accounting for resource use, and normalized cost per
success (NCS) measures the resources spent per successful task.
Across AppWorld, OfficeBench, and Tau2 Telecom with three LLM Actors,
CarryOn achieves 75.7\% macro-averaged task success, exceeding the
strongest baseline by 6.7 percentage points. It also achieves the best
average scores on both metrics.
Project repository: \url{https://anonymous.4open.science/r/CarryOn-0816/}.
\end{abstract}

\section{Introduction}
\label{sec:introduction}
\begin{figure*}[t]
    \centering
    \includegraphics[width=\textwidth]{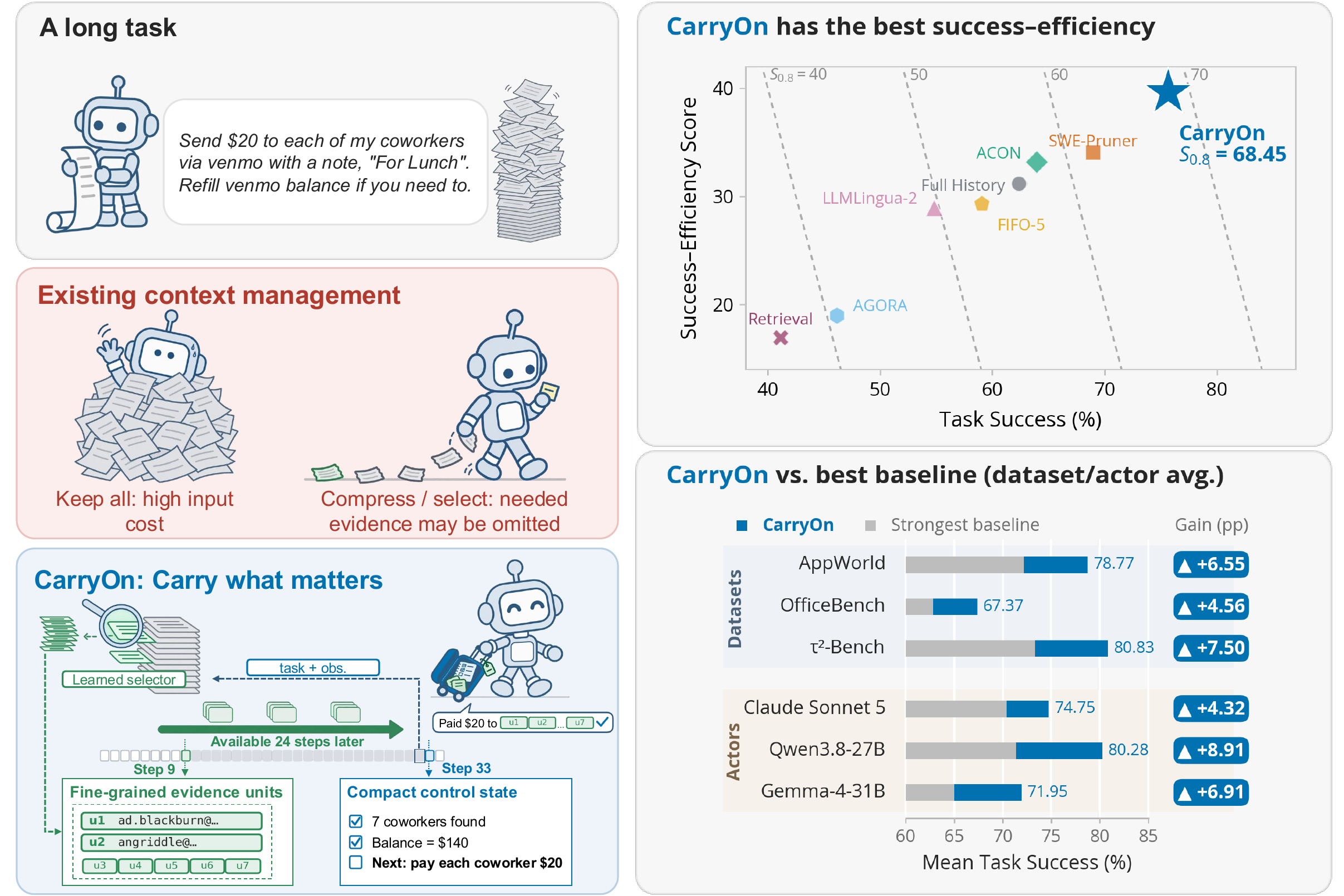}
    \input{figures/intro_caption}
\end{figure*}
Long-horizon agents built on large language models
(LLMs)~\citep{yao2022react,hu2025hiagent,wang2024agent} have made substantial
progress in solving complex, multi-step tasks. These
capabilities support applications ranging from everyday digital
tasks~\citep{trivedi2024appworld} to office
workflows~\citep{wang2024officebench} and customer
support~\citep{barres2025tau}. The payment task in the upper left of
Figure~\ref{fig:intro}, for example, requires an agent to identify
coworkers, check and refill a balance, and send payments. As such tasks
progress, observations and intermediate results accumulate, increasing
the cost of repeatedly processing history. Yet later actions may depend
on details obtained many steps earlier. This has motivated
context-management methods that limit context growth while preserving
information needed for continued
execution~\citep{hu2025hiagent,kang2025acon,zhou2026mem1}.

Some approaches fine-tune the acting LLM to manage its own memory
during task execution~\citep{zhou2026mem1,jiang2026pabu,wang2026memex}.
External context managers instead construct the Actor's input by
summarizing, pruning, or selecting interaction histories, without
fine-tuning the Actor~\citep{kang2025acon,pan2024llmlingua,zhang2026agora}.
As illustrated in the middle left of Figure~\ref{fig:intro},
retaining all history increases input cost, while compression or
selection can omit evidence needed later. Selecting complete
interactions may also retain unrelated records. The challenge is
to keep task progress concise while making the specific evidence
needed for each action available.

We propose \textbf{CarryOn}, a lightweight context manager that separates
progress tracking from evidence selection. Like carry-on luggage, it
keeps essential evidence within reach for the actions ahead.
The Actor records progress
and next steps in a compact control state, while an evidence manager
selects fine-grained units from original observations and preserves
their source references. A learned selector and adaptive retention
policy determine which evidence the Actor receives at each step,
allowing earlier records to be brought back when they become relevant.
In the lower left of Figure~\ref{fig:intro}, the
compact state tracks progress toward the payments, while coworker
records obtained at step~9 are available at step~33 to complete the
transfers after a balance refill.

We evaluate CarryOn on AppWorld, OfficeBench, and Tau2 Telecom with three
LLM Actors. Averaged equally across benchmarks and Actors, CarryOn achieves
75.7\% task success, exceeding Full History by 13.3
percentage points and the strongest overall baseline, SWE-Pruner, by
6.7 percentage points. In the upper right of Figure~\ref{fig:intro},
CarryOn achieves the highest $S_{0.8}$, a score defined in
Section~\ref{sec:metrics} that emphasizes task success while accounting
for token and Actor-call costs. CarryOn also achieves the lowest mean
normalized cost per success (NCS), at 0.792. The lower right shows higher
mean success than the strongest baseline when results are averaged by
benchmark or by Actor.

Our main contributions are:
\begin{enumerate}
    \item We introduce CarryOn, a lightweight external context manager
    that supports reliable long-horizon task execution with compact
    working context, without fine-tuning the Actor.
    \item We separate persistent progress tracking from action-specific
    evidence selection. Fine-grained original observations can be
    revisited as action requirements change, preserving task continuity
    and access to exact details without additional LLM calls for
    context management.
    \item Experiments on three benchmarks demonstrate CarryOn's
    effectiveness and cross-Actor generalization. We support this
    evaluation with two complementary metrics: $S_\alpha$ prioritizes
    success while accounting for cost, and NCS measures resource cost
    per successful outcome.
\end{enumerate}

\section{Related Work}
\label{sec:related_work}
\subsection{Long-Horizon LLM Agents}
\label{sec:related_long_horizon}
Reliable long-horizon execution requires agents to turn goals into
sequences of actions and adjust them using environment feedback.
Agents interleave reasoning and action~\citep{yao2022react} and use
executable code to compose tool calls and respond to
feedback~\citep{wang2024executable}. Reflection on previous attempts and
state-driven workflows further structure
execution~\citep{NEURIPS2023_1b44b878,wu2024stateflow}. A complementary direction reuses
successful experience as executable skills or reusable
workflows~\citep{wang2023voyager,wang2024agent}. Hierarchical working
memory further organizes ongoing execution around
subgoals~\citep{hu2025hiagent}. Evaluations that measure task length by
human completion time highlight the difficulty of sustaining reliable
execution~\citep{kwa2025measuring}. CarryOn supports long-horizon execution
by adapting the Actor's context to the information needed at each
decision step.

\subsection{Context Management for LLM Agents}
\label{sec:related_context_management}
Context management aims to keep useful information accessible within a
limited working context. One approach uses a context manager separate
from the Actor to construct its input through compression, selection,
or structured memory. Prompt compression removes
less informative content using language-model scores or learned token
classifiers~\citep{li2023compressing,jiang2023llmlingua,pan2024llmlingua}, while query-aware
compression prioritizes information relevant to the current
question~\citep{jiang2024longllmlingua,xu2024recomp}. Dialogue systems also
maintain memory by recursively summarizing previous
exchanges~\citep{wang2025recursively}. For agents, learned observation
processing and history summarization reduce the material passed to the
Actor~\citep{lee2025learning,kang2025acon}, whereas learned retention preserves selected
action--observation steps~\citep{zhang2026agora}.
Dependency-structured memory additionally connects past steps to the
earlier information they rely on, supporting selective retrieval and
summarization~\citep{wu2026contextweaver}.

Another line trains the acting LLM to manage and use memory through
recurrent consolidation, progress-aware retention, or indexed
retrieval~\citep{zhou2026mem1,jiang2026pabu,wang2026memex}.
ReSum-GRPO trains the Actor to continue from summarized
contexts~\citep{wu2025resum}.
Our work focuses on external context management. CarryOn selects
fine-grained fragments from original observations and can revisit
previously omitted evidence, while maintaining a separate compact state
for task progress.

\section{Method: CarryOn}
\label{sec:method}

CarryOn maintains an execution context that combines compact task state
with selected evidence from the interaction history. The control state
provides a concise account of progress and next steps, while selected
evidence preserves the concrete details needed for action. An evidence
manager adapts this working context to the current decision, and the
Actor uses it to act and update its state. Figure~\ref{fig:method}
organizes the method into four regions: \emph{History} (1),
\emph{Evidence selection} (2), \emph{Acting} (3), and \emph{Selector
training} (bottom). Experimental configurations appear in
Section~\ref{sec:carryon_configurations}.
\begin{figure*}[t]
    \centering
    \includegraphics[width=\textwidth]{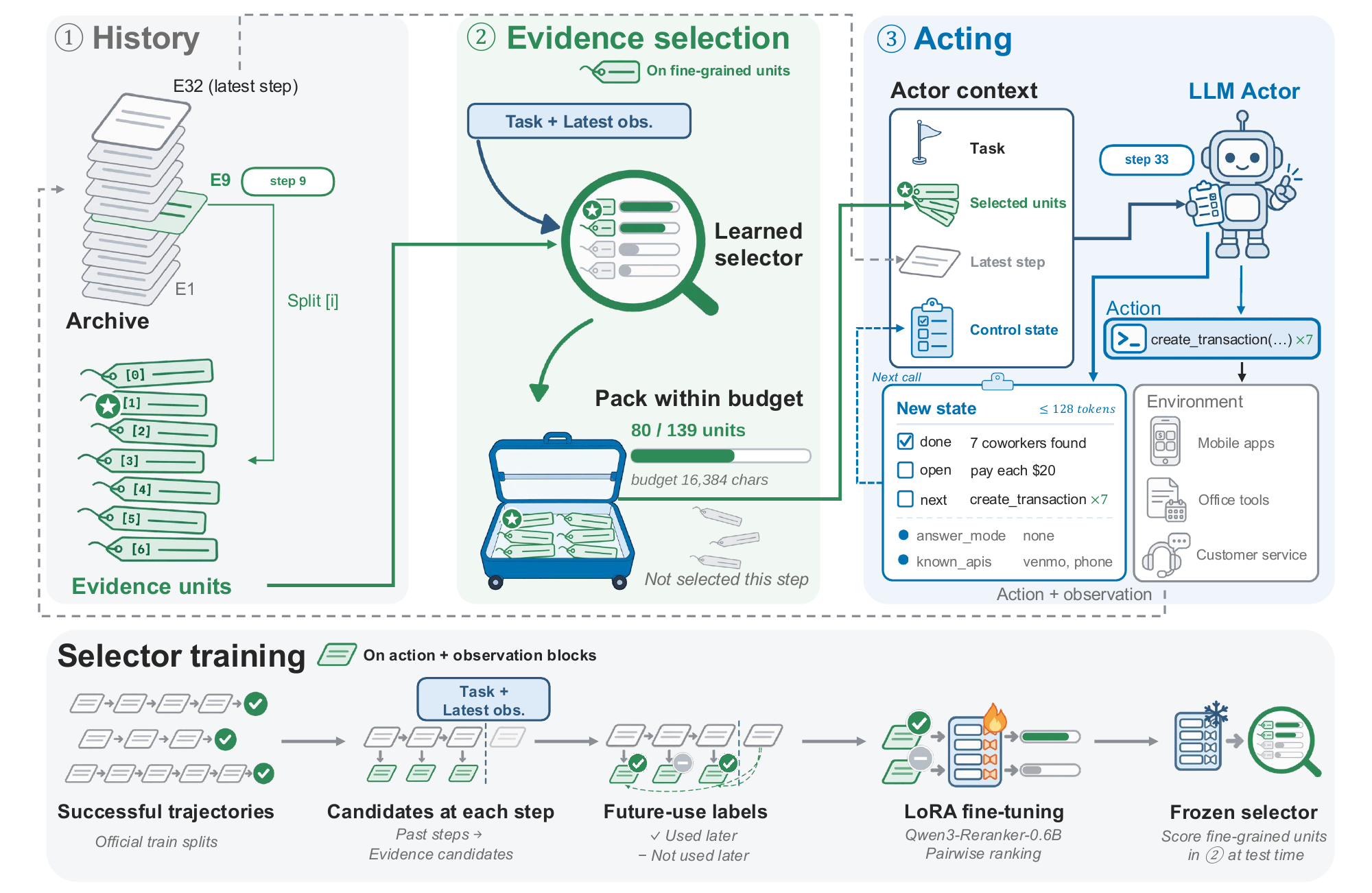}
    \input{figures/method_caption}
\end{figure*}

\subsection{Problem Formulation}
\label{sec:method_overview}
At decision step $t$, let $q$ denote the task, $H_t$ the interaction
archive, $o_t$ the current public observation, and $m_t$ the current
control state. Let $W_t$ be the selected older evidence and $R_t$ the
recent interactions always included in context. The Actor receives
\begin{equation}
    X_t = \operatorname{Compose}(q,m_t,W_t,R_t).
    \label{eq:carryon_context}
\end{equation}
Here, $\operatorname{Compose}$ also includes system instructions and
tool definitions. The Actor jointly produces an action $a_t$ and a proposed
state update $\widetilde m_{t+1}$. The accepted update carries progress
forward, while the action and its observation extend $H_t$.

\subsection{History and Evidence Units}
\label{sec:history}
Summarizing observations can omit exact details, while retaining whole
interaction steps can include many unrelated
records~\citep{kang2025acon,zhang2026agora}. CarryOn therefore selects
observation fragments independently. In region~1
(\emph{History}) of Figure~\ref{fig:method}, the \emph{Archive} stores raw
action--observation steps, and \emph{Split [i]} divides observations into
units such as individual records. Each unit retains its original content,
a reference to its source interaction, and the action that produced the
observation. Unselected evidence remains in the archive for later decisions.

\subsection{Evidence Selection}
\label{sec:learned_selection}
\label{sec:adaptive_retention}
Evidence can be essential even when its wording differs from the task.
In region~2 (\emph{Evidence selection}) of Figure~\ref{fig:method},
the \emph{Learned selector} evaluates historical units through relevance
ranking and next-action use prediction.

\paragraph{Selector scores.}
The selector uses the task $q$ and current observation $o_t$, shown as
\emph{Task + Latest obs.}, to assess each historical unit $u$.
It receives two scores:
\begin{equation}
\begin{aligned}
    r_t(u) &= f_\phi(q,o_t,u),\\
    p_t(u) &= \sigma\bigl(g_\phi(q,o_t,u)\bigr).
\end{aligned}
\label{eq:selector_scores}
\end{equation}
Here, $\phi$ denotes the selector parameters. The function $f_\phi$
produces the relevance score $r_t(u)$ for ordering candidates. The function
$g_\phi$ produces a raw next-action use score, and the sigmoid $\sigma$
maps it to $p_t(u)\in[0,1]$, the predicted probability that $u$ will be
used in the next action.

\paragraph{Temporary evidence expansion.}
Evidence needs change as a task progresses. A fragment excluded from the
base evidence pack may still be needed for the next action. Temporary
evidence expansion makes it available for the current decision, helping
the Actor proceed without repeating earlier interactions.
The \emph{Pack within budget} stage forms a base pack $P_t$ by ranking
older units with $r_t(u)$ under budget $B$, after reserving space for
recent interactions $R_t$.
The next-action use probability $p_t(u)$ then identifies additional
units to retain:
\begin{equation}
\begin{aligned}
    D_t &= \{u\in U_t\setminus P_t: p_t(u)\geq \tau\},\\
    W_t &= P_t\cup D_t,
\end{aligned}
\label{eq:next_action_expansion}
\end{equation}
where $U_t$ contains older candidates outside $R_t$, $\tau$ is the
expansion threshold, and $D_t$ contains the additional units.
This expansion can exceed the base budget and is recomputed at each
decision step as evidence needs change. The final evidence $W_t$ follows the green arrow into
region~3 (\emph{Acting}) of Figure~\ref{fig:method}.
Appendix~\ref{sec:carryon_runtime_details} gives the budget accounting
and full algorithm.

\subsection{Acting with State and Evidence}
\label{sec:acting}
\label{sec:state_evidence}
A compact state tracks what to do next, while evidence supplies the
concrete values needed to act, such as account identifiers, file paths,
or user-provided facts (Appendix~\ref{sec:cross_environment_examples}).
In the payment example in region~3 (\emph{Acting}) of
Figure~\ref{fig:method}, the state records that seven coworkers have been
found and should each receive \$20, while evidence supplies the recipients'
account identifiers.

\paragraph{Context and state update.}
The \emph{Actor context} brings these representations together with the
task and \emph{Latest step} (Eq.~\ref{eq:carryon_context}). The
\emph{LLM Actor} jointly produces \emph{Action} and \emph{New state}.
In our implementation, the
Actor's bounded \textbf{\texttt{MEMORY}} update has five fields:
\textbf{\texttt{done}} records completed work,
\textbf{\texttt{open}} remaining obligations,
\textbf{\texttt{next}} the next operation,
\textbf{\texttt{answer\_mode}} whether a returned value is required, and
\textbf{\texttt{known\_apis}} previously inspected APIs.
The blue \emph{Next call} arrow carries the accepted state into the next
decision. Executing the action in the \emph{Environment} adds new evidence
through the \emph{Action + observation} return path.

\paragraph{Adaptive recent retention.}
Recent interactions provide the context needed to continue execution and
correct invalid responses. CarryOn always retains the latest interaction
and extends this recent window when the Actor produces an invalid action
or state update. This adjusts the \emph{Latest step} input in region~3 of
Figure~\ref{fig:method}. Interactions outside this window remain candidates for selection.
Appendix~\ref{sec:carryon_runtime_details} specifies the acceptance and
retention rules.

\subsection{Selector Training}
\label{sec:selector_training}
Successful trajectories show how earlier evidence supports later actions.
The bottom \emph{Selector training} region of Figure~\ref{fig:method}
uses these trajectories from official training splits to supervise
selection. At the \emph{Future-use labels} stage, evidence cited in state
annotations or used in later actions provides positive ranking examples.
Candidates with neither signal serve as weak negatives. Next-action use
labels identify evidence used in the immediately following action.

The \emph{LoRA fine-tuning} stage~\citep{hu2021lora} first adapts a
pretrained reranker for pairwise ranking. For a positive block $u^+$
and a weak negative block $u^-$ from the same decision step, the loss is
\begin{equation}
    \ell_{\mathrm{rank}}
    = \log\!\left(1+\exp(\gamma+r^- -r^+)\right),
    \label{eq:rank_loss}
\end{equation}
where $r^+=r_t(u^+)$ and $r^-=r_t(u^-)$ are their relevance scores, and
$\gamma$ is the ranking margin.
We then jointly optimize this ranking objective and binary cross-entropy
for next-action use prediction:
\begin{equation}
    \ell_{\mathrm{use}}=-y\log p-(1-y)\log(1-p),
    \label{eq:next_action_loss}
\end{equation}
where $p=p_t(u)$ and $y\in\{0,1\}$ labels whether block $u$ is used in
the immediately following action.
Both stages use complete action--observation blocks, matching the evidence
references in the annotations. At test time, the \emph{Frozen selector}
scores individual fragments in region~2.
Appendix~\ref{sec:selector_training_details} gives the labeling and
optimization details.

\subsection{CarryOn-E: Local Deployment}
\label{sec:method_local_actors}
As a local deployment extension, CarryOn-E distills a 27B CarryOn Actor
into a 9B student while keeping the context manager fixed, with the
training procedure and results in Appendix~\ref{sec:local_distillation}.

\section{Experimental Setup}
\label{sec:experimental_setup}

\subsection{Datasets and Baselines}
\label{sec:benchmarks}
\label{sec:baseline_families}
We evaluate CarryOn on AppWorld~\citep{trivedi2024appworld},
OfficeBench~\citep{wang2024officebench}, and the Telecom domain of
Tau2-Bench~\citep{barres2025tau}, covering cross-application tasks,
office workflows, and conversational customer support. The evaluation
sets contain 168, 95, and 40 tasks, respectively, with success determined
by each benchmark's official evaluator.

We compare against Full History, FIFO-5, and Retrieval, as well as
LLMLingua-2~\citep{pan2024llmlingua}, ACON~\citep{kang2025acon}, and
SWE-Pruner~\citep{wang2026swe} for compression and pruning.
AGORA~\citep{zhang2026agora} provides a learned history-selection baseline.
Dataset splits and baseline configurations are given in
Appendix~\ref{sec:experimental_details}.

\subsection{Experimental Configuration}
\label{sec:actors}
\label{sec:carryon_configurations}
We evaluate Claude Sonnet 5,\footnote{\url{https://www.anthropic.com/news/claude-sonnet-5}}
Qwen3.8-27B,\footnote{\url{https://huggingface.co/Qwen/Qwen3.8-27B}}
and Gemma-4-31B.\footnote{\url{https://huggingface.co/google/gemma-4-31B-it}}
For each benchmark, we reuse the same trained CarryOn selector across
all three Actors. ACON transfers prompts optimized with Claude as the
Actor. AGORA transfers Claude-labeled scorers to Qwen and Gemma on all
three benchmarks.
LLMLingua-2 and SWE-Pruner, like CarryOn, use external models without
Actor-specific retraining. The CarryOn selector uses Qwen3-Reranker-0.6B, with offline
state annotations and evidence references generated by Claude Sonnet 5.
Context budgets, training settings, and execution interfaces are
detailed in Appendix~\ref{sec:experimental_details}.

\subsection{Evaluation Metrics}
\label{sec:metrics}
Context management should support successful task completion while
controlling execution cost. Success alone does not reveal the resources
required~\citep{kapoor2024ai}, while low total cost can reflect early
failure. We therefore
report Success with two complementary measures: $S_\alpha$ emphasizes
task capability while accounting for cost, and normalized cost per
success (NCS) measures the resources spent per successful outcome.
Task Goal Completion (TGC), reported as Success (\%) in the tables, is
$100s/N$, where $s$ is the number of successful tasks among all $N$
evaluation tasks. All cost measures include successful and failed tasks.
Tokens count submitted inputs and returned outputs, including inputs
rejected for exceeding the context limit.

\paragraph{Success--efficiency score.}
Motivated by joint quality--efficiency evaluation~\citep{chi2026agentmeter},
we define $S_\alpha$ to emphasize task success while rewarding efficient
execution. We first discount success by normalized mean token and
Actor-call costs, then combine this score with Success:
\begin{align}
S_0 &= \frac{\mathrm{TGC}}
 {1+\tfrac{1}{2}\bar{T}/B_T+\tfrac{1}{2}\bar{C}/B_C},
\label{eq:efficiency_score}\\
S_\alpha &= \alpha\,\mathrm{TGC}
 +(1-\alpha)\,S_0.
\label{eq:success_efficiency_score}
\end{align}
Here $\bar{T}$ and $\bar{C}$ are mean system request tokens and Actor
calls over all $N$ tasks. The reference values $B_T$ and $B_C$ are the
corresponding Full History means on the same task set.
We report $S_{0.8}$, assigning 80\% weight to Success and 20\% to its
cost-discounted counterpart. This keeps task completion central while
rewarding lower resource use. Higher scores are better.

\paragraph{Normalized cost per success.}
Averaging costs only over successful trajectories omits the resources
spent on failures. Following the cost-of-pass
perspective~\citep{wang2025efficient}, we define NCS by dividing total
costs by the number of successes and normalizing these ratios against
Full History (FH):
\begin{equation}
\mathrm{NCS}=
\frac{1}{2}\frac{C/s}{C^{\mathrm{FH}}/s^{\mathrm{FH}}}
+\frac{1}{2}\frac{T/s}{T^{\mathrm{FH}}/s^{\mathrm{FH}}}.
\label{eq:ncs}
\end{equation}
Here $C$ and $T$ are total Actor calls and system request tokens over
all $N$ tasks, including failures, and $s$ is the successful-task count.
NCS equally weights relative calls and tokens per success. Lower values
are better, with Full History equal to 1. NCS emphasizes resource
efficiency and is interpreted alongside Success.

Scores are calculated within each dataset--Actor setting before
averaging settings equally. Token components and additional accounting
details appear in
Appendix~\ref{sec:cost_accounting_details}.

\section{Results and Analysis}
\label{sec:analysis}

\subsection{Main Results}
\label{sec:main_results}
\label{sec:capability_efficiency}
Table~\ref{tab:main_results} compares CarryOn with seven baselines across
three benchmarks and three Actors. CarryOn achieves the highest mean
Success (75.7\%) and $S_{0.8}$ (68.45), together with the lowest mean NCS
(0.792). It ranks first in Success in seven comparisons and ties for
first in one, achieves the highest $S_{0.8}$ in eight, and has the lowest
NCS in six. The same frozen selector is reused within each benchmark
(Section~\ref{sec:selector_training}). \textbf{Separating evidence
selection from action generation enables cross-Actor reuse:} one
frozen selector supports gains across different Actors.
CarryOn also retains the highest mean $S_\alpha$ and lowest mean NCS
as their respective weights vary over $[0,1]$
(Appendix~\ref{sec:metric_weight_sensitivity}).

\begin{table*}[t]
\centering
\small
\setlength{\tabcolsep}{5pt}
\renewcommand{\arraystretch}{1.12}
\begin{tabular}{ll*{9}{r}}
\toprule
Dataset & Method & \multicolumn{3}{c}{Claude Sonnet 5} & \multicolumn{3}{c}{Qwen3.8-27B} & \multicolumn{3}{c}{Gemma-4-31B} \\
\cmidrule(lr){3-5}\cmidrule(lr){6-8}\cmidrule(lr){9-11}
 & & \shortstack{Success\\(\%) $\uparrow$} & \shortstack{$S_{0.8}$\\$\uparrow$} & \shortstack{NCS\\$\downarrow$} & \shortstack{Success\\(\%) $\uparrow$} & \shortstack{$S_{0.8}$\\$\uparrow$} & \shortstack{NCS\\$\downarrow$} & \shortstack{Success\\(\%) $\uparrow$} & \shortstack{$S_{0.8}$\\$\uparrow$} & \shortstack{NCS\\$\downarrow$} \\
\midrule
 & Full History & 43.5 & 39.11 & 1.000 & 65.5 & 58.93 & \textbf{1.000} & 53.6 & 48.21 & 1.000 \\
\cmidrule(l){2-11}
 & FIFO-5 & 42.9 & 37.62 & 1.592 & 53.6 & 47.83 & 1.410 & 48.8 & 43.96 & 1.084 \\
 & Retrieval & 11.9 & 10.24 & 8.549 & 44.0 & 38.50 & 2.525 & 28.6 & 24.80 & 3.636 \\
\raisebox{-.5\baselineskip}[0pt][0pt]{AppWorld} & LLMLingua-2 & 21.4 & 19.20 & 2.195 & 44.6 & 39.90 & 1.658 & 41.7 & 37.77 & 1.127 \\
 & ACON & 70.2 & 62.84 & 0.688 & 53.0 & 47.08 & 1.553 & 48.8 & 44.01 & 1.063 \\
 & AGORA & 55.4 & 48.31 & 1.376 & 41.1 & 36.65 & 1.862 & 39.3 & 34.71 & 1.899 \\
 & SWE-Pruner & 79.2 & 70.57 & 0.652 & 68.5 & 59.24 & 1.970 & 69.0 & 62.17 & \textbf{0.769} \\
\cmidrule(l){2-11}
\rowcolor{black!6}
\cellcolor{white} & \textbf{CarryOn} & \textbf{79.8} & \textbf{71.71} & \textbf{0.556} & \textbf{84.5} & \textbf{73.06} & 1.634 & \textbf{72.0} & \textbf{64.20} & 0.885 \\
\midrule
 & Full History & 63.2 & 56.84 & 1.000 & \textbf{66.3} & 59.68 & 1.000 & 51.6 & 46.42 & 1.000 \\
\cmidrule(l){2-11}
 & FIFO-5 & 57.9 & 53.17 & 0.751 & 53.7 & 48.41 & 1.191 & 54.7 & 49.96 & 0.728 \\
 & Retrieval & 49.5 & 44.67 & 1.207 & 51.6 & 45.87 & 1.592 & 34.7 & 31.14 & 1.593 \\
\raisebox{-.5\baselineskip}[0pt][0pt]{OfficeBench} & LLMLingua-2 & 64.2 & 59.22 & 0.625 & 60.0 & 54.48 & 0.942 & 53.7 & 49.66 & 0.577 \\
 & ACON & 67.4 & 61.96 & 0.629 & 63.2 & 57.54 & 0.841 & 57.9 & 53.40 & 0.565 \\
 & AGORA & 49.5 & 44.55 & 1.263 & 46.3 & 41.31 & 1.682 & 38.9 & 34.87 & 1.453 \\
 & SWE-Pruner & 62.1 & 57.03 & 0.703 & 63.2 & 56.61 & 1.129 & 61.1 & 56.14 & 0.569 \\
\cmidrule(l){2-11}
\rowcolor{black!6}
\cellcolor{white} & \textbf{CarryOn} & \textbf{69.5} & \textbf{64.36} & \textbf{0.530} & \textbf{66.3} & \textbf{62.04} & \textbf{0.476} & \textbf{66.3} & \textbf{62.30} & \textbf{0.338} \\
\midrule
 & Full History & 75.0 & 67.50 & 1.000 & 80.0 & 72.00 & 1.000 & 62.5 & 56.25 & 1.000 \\
\cmidrule(l){2-11}
 & FIFO-5 & 72.5 & 64.61 & 1.234 & 87.5 & 78.46 & 0.976 & 60.0 & 53.96 & 1.056 \\
 & Retrieval & 42.5 & 36.58 & 4.050 & 72.5 & 64.49 & 1.361 & 35.0 & 30.42 & 3.375 \\
\raisebox{-.5\baselineskip}[0pt][0pt]{Tau2 Telecom} & LLMLingua-2 & \textbf{80.0} & \textbf{71.81} & \textbf{0.982} & 80.0 & 72.00 & 0.999 & 47.5 & 42.57 & 1.419 \\
 & ACON & 72.5 & 65.41 & 0.991 & 87.5 & 79.03 & 0.858 & 55.0 & 48.87 & 1.433 \\
 & AGORA & 62.5 & 54.90 & 1.859 & 50.0 & 42.91 & 3.896 & 32.5 & 28.34 & 3.417 \\
 & SWE-Pruner & 70.0 & 63.06 & 1.053 & 82.5 & 74.05 & 1.018 & 65.0 & 58.85 & 0.864 \\
\cmidrule(l){2-11}
\rowcolor{black!6}
\cellcolor{white} & \textbf{CarryOn} & 75.0 & 67.07 & 1.121 & \textbf{90.0} & \textbf{81.46} & \textbf{0.803} & \textbf{77.5} & \textbf{69.87} & \textbf{0.782} \\
\bottomrule
\end{tabular}
\caption{Main results across three benchmarks and three Actors. Bold marks the best result for each benchmark and Actor, including ties.}
\label{tab:main_results}
\end{table*}

On AppWorld, CarryOn leads in Success and $S_{0.8}$ for all three
Actors. The largest improvement over the strongest baseline occurs
with Qwen, where Success rises from SWE-Pruner's 68.5\% to 84.5\%.
This is consistent with the state--evidence division in
Section~\ref{sec:state_evidence}: compact state tracks unfinished work,
while selected evidence supplies exact values for tool calls.
The drop in Success when state is removed
(Section~\ref{sec:component_ablations}) supports the role of progress
tracking alongside evidence selection. Section~\ref{sec:evidence_analysis}
shows how earlier records remain available for later actions.

OfficeBench shows the clearest gains in resource efficiency. CarryOn
achieves the highest $S_{0.8}$ and lowest NCS for every Actor, while
leading in Success for Claude and Gemma and matching Full History for
Qwen. Both methods complete 66.3\% of Qwen tasks, but CarryOn raises $S_{0.8}$
from 59.68 to 62.04 and lowers NCS from 1.000 to 0.476.
This is the intended benefit of selecting individual observation
fragments (Section~\ref{sec:history}): limiting repeated history while
keeping the source records needed for ongoing work. Compact state
preserves progress as the selected evidence changes
(Section~\ref{sec:state_evidence}), supporting continued execution
with a smaller working context.

On Tau2 Telecom, CarryOn leads all three metrics with Qwen and Gemma.
For Gemma, Success reaches 77.5\%, compared with 65.0\% for the
strongest baseline, SWE-Pruner, while NCS falls from 0.864 to 0.782.
Here the state--evidence division supports dialogue: state tracks
pending diagnostic steps, while evidence preserves customer reports
and tool-returned account facts. The recorded example in
Appendix~\ref{sec:cross_environment_examples} shows these two inputs
guiding the next response.
With Claude, LLMLingua-2 achieves higher Success (80.0\% vs.\ 75.0\%).

Gains also appear across different task groups. Averaged across Actors,
CarryOn exceeds the strongest baseline within each group by 12.28 and
9.72 percentage points on AppWorld easy and medium tasks, 8.60 points on
OfficeBench L3, and 6.25 points on Tau2 MMS tasks. Performance in the
remaining groups is broadly comparable to their strongest baselines,
with the complete breakdown in Appendix~\ref{sec:stratified_results}.

\begin{figure*}[t]
    \centering
    \includegraphics[width=0.95\textwidth]{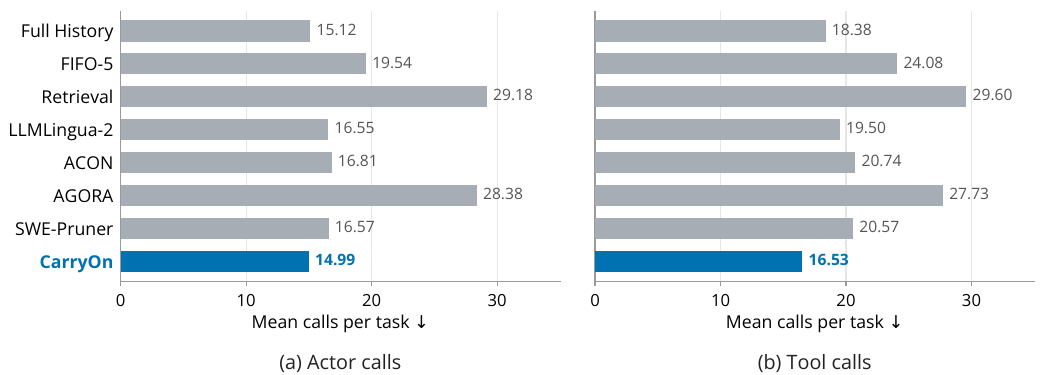}
    \caption{Mean calls per task, averaged equally across benchmarks and Actors.
    Failed tasks are included.}
    \label{fig:call_efficiency}
\end{figure*}

\subsection{Evidence Retention for Long-Term Dependencies}
\label{sec:evidence_analysis}
\begin{figure}[t]
    \centering
    \includegraphics[width=\columnwidth]{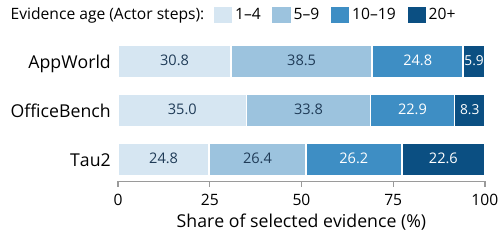}
    \caption{Age of evidence selected by CarryOn, excluding automatic recent retention.
    Percentages pool selection events across Actors within each benchmark.}
    \label{fig:evidence_reuse}
\end{figure}

The ranking objective in Eq.~\ref{eq:rank_loss} teaches the selector to
favor evidence cited in state annotations or used in later actions.
At each decision, CarryOn recomputes the relevance score $r_t(u)$ using
the task and current observation (Section~\ref{sec:learned_selection}),
allowing previously omitted evidence to be selected again.
Figure~\ref{fig:evidence_reuse} shows that evidence obtained at least
ten Actor steps earlier accounts for 30.7\% of selections in AppWorld,
31.2\% in OfficeBench, and 48.8\% in Tau2 Telecom. In 40.2\% of task
runs, evidence is selected again after being omitted for one or more
steps. The payment case in Appendix~\ref{sec:evidence_reuse_details}
illustrates the role of relevance ranking: a previously omitted card
record returns to the base evidence pack, and the Actor uses its details
to add the card and refill the balance. \textbf{Relevance ranking brings
earlier evidence back when action requirements change.}

Temporary evidence expansion adds records predicted to be used in the
next action (Eq.~\ref{eq:next_action_expansion}). It supplements the base
pack in 4.4\% of Actor steps on AppWorld and 12.2\% on OfficeBench.
In an AppWorld example, expansion restores an omitted contact record,
and the Actor immediately uses its phone number to send a message
(Appendix~\ref{sec:expansion_case}).

\subsection{More Efficient Agent Trajectories}
\label{sec:trajectory_efficiency}
\textbf{CarryOn completes more tasks with fewer tool calls on average.}
It has the lowest mean Actor and tool calls
(Figure~\ref{fig:call_efficiency}), using 10.1\% fewer tool calls than
Full History with comparable Actor calls and higher Success
(Table~\ref{tab:main_results}), consistent with state tracking pending
work and selection enabling evidence reuse
(Section~\ref{sec:state_evidence}).

\subsection{Key Component Ablations}
\label{sec:component_ablations}
Table~\ref{tab:appworld_claude_ablations} compares four variants on
AppWorld with Claude. \textbf{w/o MEMORY} removes control state and
reduces Success most, from 79.8\% to 66.1\%.
\textbf{Ranking only} removes next-action training and
temporary evidence expansion, yielding 76.8\%. \textbf{Fine-grained
training} uses individual units labeled by later API-call reuse,
without state-annotation citations, and achieves 73.2\%.
\textbf{Recency packing} replaces learned selection and temporary
expansion with the most recent units under the same base budget,
yielding 76.2\%. \textbf{CarryOn} leads all three metrics.
\textbf{Compact state and learned evidence selection are complementary,}
outperforming either state removal or recency-based selection.

\begin{table}[t]
\centering
\small
\setlength{\tabcolsep}{3pt}
\renewcommand{\arraystretch}{1.15}
\begin{tabular*}{\columnwidth}{@{\extracolsep{\fill}}lrrr@{}}
\toprule
Method & Success (\%) $\uparrow$ & $S_{0.8}$ $\uparrow$ & NCS $\downarrow$ \\
\midrule
w/o MEMORY & 66.1 & 58.72 & 0.824 \\
Ranking only & 76.8 & 69.09 & 0.568 \\
Fine-grained training & 73.2 & 65.68 & 0.629 \\
Recency packing & 76.2 & 68.36 & 0.603 \\
\midrule
\textbf{CarryOn} & \textbf{79.8} & \textbf{71.71} & \textbf{0.556} \\
\bottomrule
\end{tabular*}
\caption{Component and training variants on AppWorld with Claude Sonnet 5.
Bold marks the best result.}
\label{tab:appworld_claude_ablations}
\end{table}

\subsection{MEMORY Capacity Sensitivity}
\label{sec:capacity_analysis}
On AppWorld, capacities of 64, 128, and 256 proxy tokens yield 130,
134, and 131 successes, respectively (Table~\ref{tab:memory_capacity}).
The 128-token setting achieves the highest Success and $S_{0.8}$,
while 256 tokens achieve the lowest NCS. The smaller 64-token state
is truncated more often. \textbf{A larger state does not consistently
improve task success.} Appendix~\ref{sec:capacity_details} provides
the full results.

\section{Conclusion}
\label{sec:conclusion}
CarryOn separates compact progress tracking from fine-grained evidence
selection, keeping exact observations available as action requirements
change. Experiments across three benchmarks and three Actors demonstrate
its effectiveness and cross-Actor generalization. CarryOn achieves 75.7\%
mean task success, exceeding the strongest baseline by 6.7 percentage
points, and leads both $S_{0.8}$ and NCS on average. These results
support adapting evidence to each decision while maintaining concise
task state for long-horizon execution.

\section*{Limitations}
\label{sec:limitations}
Our evaluation covers three benchmarks with text-based interactions.
Extending CarryOn to visual or multimodal observations would require
adapting how evidence is represented and selected. We train a separate
selector for each benchmark and reuse it across Actors. Generalization
to new environments, particularly those with different tool interfaces
and observation structures, remains to be explored.

In real-world applications, retained observations may contain sensitive
information, while missing or misinterpreted evidence may lead to
incorrect actions. Deployment therefore requires appropriate access
controls for stored history and user confirmation for consequential
actions.

\bibliography{carryon_references}
\clearpage

\appendix
\raggedbottom
\section{Experimental Details}
\label{sec:experimental_details}
This appendix provides the dataset splits, configuration settings, and
execution interfaces summarized in Section~\ref{sec:experimental_setup}.

\subsection{Benchmarks and Task Splits}
\label{sec:benchmark_details}
AppWorld~\citep{trivedi2024appworld} uses the official
\texttt{test\_normal} split of 168 tasks, comprising 57 easy, 48 medium,
and 63 hard tasks. OfficeBench~\citep{wang2024officebench} uses the
ACON-public test set of 95 tasks, comprising 42 L1, 22 L2, and 31 L3
tasks with 209 planned evaluator conditions. Tau2
Telecom~\citep{barres2025tau} uses the official 40-task test set.
We retain the benchmarks' native tasks, tools, and official success
criteria. CarryOn selector training uses successful trajectories from
the official training splits, as described in
Appendix~\ref{sec:selector_training_details}.

\subsection{CarryOn and Execution Configuration}
\label{sec:carryon_configuration_details}
\paragraph{Models and computation.}
Each local training job uses one NVIDIA H100 GPU. The selector has
0.6B parameters, while the Qwen and Gemma Actors have 27B and 31B,
respectively; CarryOn-E uses a 9B student. For the reported selectors,
the two training stages together take approximately 4.87, 0.59, and
0.79 hours on AppWorld, OfficeBench, and Tau2, respectively, estimated
from training logs. These are single-GPU wall times, excluding
trajectory generation and evaluation.
We report a single run per setting.

\paragraph{Context construction.}
We use a base evidence budget of $B=16{,}384$ characters and a temporary
evidence expansion threshold of $\tau=0.5$. The MEMORY capacity is 128
proxy tokens, counted locally as words and punctuation symbols.
The recent-evidence window retains the newest interaction after an
accepted response and the two newest otherwise, limited to available
history. Handling of public interaction boundaries follows
Appendix~\ref{sec:carryon_runtime_details}.
The budget covers the base evidence pack.

\paragraph{Selector training.}
The selector backbone is Qwen3-Reranker-0.6B. We use Claude Sonnet 5 to
generate the offline state annotations and their evidence references.
The pairwise ranking margin is $\gamma=0.2$.
Training objectives and label construction are specified in
Appendix~\ref{sec:selector_training_details}.

\paragraph{Execution interfaces.}
Actors use benchmark-specific action and output interfaces. Tau2 pairs
the Actor with a Haiku 4.5 User simulator. The Actor and selector receive
only public interactions, excluding private simulator state.

\subsection{Baseline Configurations}
\label{sec:baseline_configurations}
We retain each method's context construction and evaluate the resulting
systems on the same task sets. Cost accounting is described in
Appendix~\ref{sec:cost_accounting_details}.

\paragraph{Simple history baselines.}
Full History retains the complete interaction history. FIFO-5 retains
the five most recent action--observation pairs. Retrieval selects four
historical interactions using \texttt{text-embedding-3-large} and includes
the latest interaction. FIFO-5 and Retrieval follow the implementations
in ACON~\citep{kang2025acon}.

\paragraph{LLMLingua-2.}
We use LLMLingua-2~\citep{pan2024llmlingua} as a local history compressor
with a fixed keep rate of 0.3.

\paragraph{ACON and AGORA.}
ACON~\citep{kang2025acon} uses LLM-based context compression.
AGORA~\citep{zhang2026agora} uses a learned scorer to select historical
interactions.

\paragraph{SWE-Pruner.}
We use benchmark-adapted SWE-Pruner models~\citep{wang2026swe} with a
pruning threshold of 0.4. Observations are pruned when the Actor provides
a focus and otherwise retained.

\section{Execution and Selector Training Details}
\label{sec:method_details}
This appendix expands the execution and training procedures in
Section~\ref{sec:method}.

\subsection{Context Assembly and Recent Retention}
\label{sec:carryon_runtime_details}
\paragraph{Base budget accounting.}
With base evidence budget $B$ and rendered character count $c(\cdot)$,
recent interactions $R_t$ leave the following capacity for older units:
\begin{equation}
    B_t^{\mathrm{old}}=\max\{0,B-c(R_t)\}.
    \label{eq:evidence_budget}
\end{equation}
After reserving space for recent interactions, CarryOn selects older
evidence in order of relevance, skipping units that exceed the remaining
budget. If several selected fragments come from the same observation,
the action that produced it is included only once. Recent interactions
are always retained, even when they exceed the base budget. Temporary
expansion can add further evidence. CarryOn rebuilds the evidence pack
at each step to reflect changing needs.

\paragraph{Adaptive recent retention.}
CarryOn always retains the latest interaction. At the next decision, it
extends the recent window if the action fails validation or the MEMORY
update is invalid or truncated. Otherwise, it uses the standard window.
Unparseable updates preserve the previous state. Interactions leaving
the window remain available for learned selection. Window lengths are
given in Appendix~\ref{sec:carryon_configuration_details}.

\paragraph{Execution loop.}
Algorithm~\ref{alg:carryon} summarizes the three upper regions of
Figure~\ref{fig:method}. Valid state updates are bounded by the state
capacity, while invalid updates preserve the previous state.
\begin{algorithm}[t]
\caption{CarryOn execution loop}
\label{alg:carryon}
\small
\begin{algorithmic}[1]
\Require Task $q$, frozen selector $\phi$, budget $B$, threshold $\tau$
\State Initialize archive $H_0$, state $m_0$, observation $o_0$
\State $t\gets 0$
\While{task execution continues}
    \State $R_t\gets$ recent interactions under adaptive retention
    \State $U_t\gets$ evidence units from $H_t$ outside $R_t$
    \State Compute $r_t(u),p_t(u)$ for $u\in U_t$
    \Statex \hspace{\algorithmicindent}\hspace{\algorithmicindent}using $q,o_t$ (Eq.~\ref{eq:selector_scores})
    \State $B_t^{\mathrm{old}}\gets\max\{0,B-c(R_t)\}$
    \State $P_t\gets$ units selected by decreasing $r_t(u)$
    \Statex \hspace{\algorithmicindent}\hspace{\algorithmicindent}within $B_t^{\mathrm{old}}$, skipping units that do not fit
    \State $D_t\gets\{u\in U_t\setminus P_t:p_t(u)\geq\tau\}$
    \State $W_t\gets P_t\cup D_t$
    \State $X_t\gets\operatorname{Compose}(q,m_t,W_t,R_t)$
    \State $(a_t,\widetilde m_{t+1})\gets\operatorname{Actor}(X_t)$
    \State Set $m_{t+1}$ to the bounded update, or $m_t$ if invalid
    \State Submit $a_t$ and observe public feedback $o_{t+1}$
    \State Extend $H_t$ with $(a_t,o_{t+1})$ to form $H_{t+1}$
    \State $t\gets t+1$
\EndWhile
\end{algorithmic}
\end{algorithm}

\subsection{Selector Supervision and Optimization}
\label{sec:selector_training_details}
\paragraph{Training examples and labels.}
We construct examples at intermediate decision boundaries of successful
trajectories from the official training splits. In the block-supervised
path, each candidate is an earlier complete action--observation block.
Ranking positives combine evidence references in offline state annotations
with observed evidence-use signals. AppWorld traces which earlier
observations supply values to the next tool call. OfficeBench matches
earlier evidence to arguments of the next action. Tau2 tracks evidence
reuse in subsequent public interactions.
For next-action use prediction, positive labels are restricted to evidence
linked to the immediately following action.

Uncited and unused evidence
is treated as a weak negative. Offline labels use subsequent public
interactions, while selector inputs contain the task and observations
available at the current decision.

\paragraph{Scoring architecture and objectives.}
The functions $f_\phi$ and $g_\phi$ in Eq.~\ref{eq:selector_scores} share
a pretrained backbone with separate instructions for relevance
ranking and next-action use prediction. The latter uses a learned linear
head. The sigmoid is $\sigma(z)=1/(1+\exp(-z))$.
The initial training stage adapts the reranker with LoRA using the pairwise ranking loss
in Eq.~\ref{eq:rank_loss}.
The backbone and margin are specified in
Appendix~\ref{sec:carryon_configuration_details}.
The subsequent stage jointly optimizes this ranking objective
and the next-action use loss in Eq.~\ref{eq:next_action_loss}. This stage updates
the ranker adapter as well as the new head.

Whole-block training preserves the units referenced by the offline
annotations. At deployment, the same query and candidate interface scores
the finer units from Section~\ref{sec:history}, each paired with the action
that produced its source observation. This transfers block-level
supervision to fragment selection without requiring a separate annotation
for every fragment.

\section{MEMORY Capacity Sensitivity}
\label{sec:capacity_details}
\begin{figure*}[t]
\centering
\includegraphics[width=\textwidth]{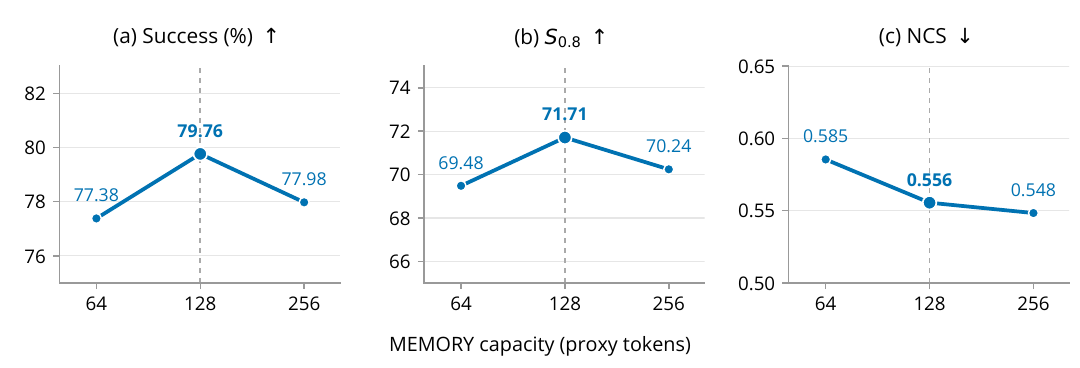}
\caption{MEMORY capacity sensitivity on AppWorld with Claude.
Dashed lines mark the default capacity of 128 proxy tokens.}
\label{fig:memory_capacity}
\end{figure*}

Figure~\ref{fig:memory_capacity} and Table~\ref{tab:memory_capacity}
compare control-state capacities of 64,
128, and 256 proxy tokens on AppWorld with Claude. Each setting is
evaluated once on all 168 tasks, with the evidence budget, selector,
and retention policy held fixed.

The default capacity of 128 tokens achieves the highest Success and
$S_{0.8}$. Increasing capacity to 256 gives the lowest NCS and reduces
calls and token use, while
the 64-token state is truncated more often. The results illustrate how
state capacity affects both task completion and resource use.

\begin{table*}[t]
\centering
\small
\setlength{\tabcolsep}{4pt}
\renewcommand{\arraystretch}{1.12}
\begin{tabular}{lrrrrrr}
\toprule
Capacity & Success (\%) $\uparrow$ & $S_{0.8}$ $\uparrow$ & NCS $\downarrow$ & Calls & Tokens (M) & Truncations \\
\midrule
64 & 77.4 & 69.48 & 0.585 & 3,163 & 25.507 & 284 \\
128 & \textbf{79.8} & \textbf{71.71} & 0.556 & 3,098 & 24.906 & 48 \\
256 & 78.0 & 70.24 & \textbf{0.548} & 3,001 & 23.920 & 8 \\
\bottomrule
\end{tabular}
\caption{MEMORY capacity sensitivity on AppWorld with Claude. Capacity is measured in proxy tokens.}
\label{tab:memory_capacity}
\end{table*}

\section{Detailed Results and Resource Use}
\label{sec:detailed_accounting}
The following tables expand Table~\ref{tab:main_results}, with results
grouped by benchmark and Actor.

\subsection{Evaluation and Cost Accounting}
\label{sec:cost_accounting_details}
TGC and Success report completed tasks out of all evaluated tasks.
Requirements reports passed checks out of all task requirements in
AppWorld. Conditions reports passed checks out of the conditions evaluated
in OfficeBench. Tau2's average reward equals its success rate under
single-run evaluation.

Calls reports Actor calls in AppWorld and OfficeBench. For Tau2, A/U/C
denotes Actor, User, and compressor calls. Token counts are reported in
millions. AppWorld and OfficeBench report Actor tokens under Input and
Output, while Total also includes compression and retrieval-embedding
requests. Tau2 token counts additionally include the User simulator.
Tokens processed by the local models used for selection or pruning in
CarryOn, AGORA, LLMLingua-2, and SWE-Pruner are excluded.

\subsection{Metric Weight Sensitivity}
\label{sec:metric_weight_sensitivity}
\begin{figure*}[!tp]
    \centering
    \includegraphics[width=\textwidth]{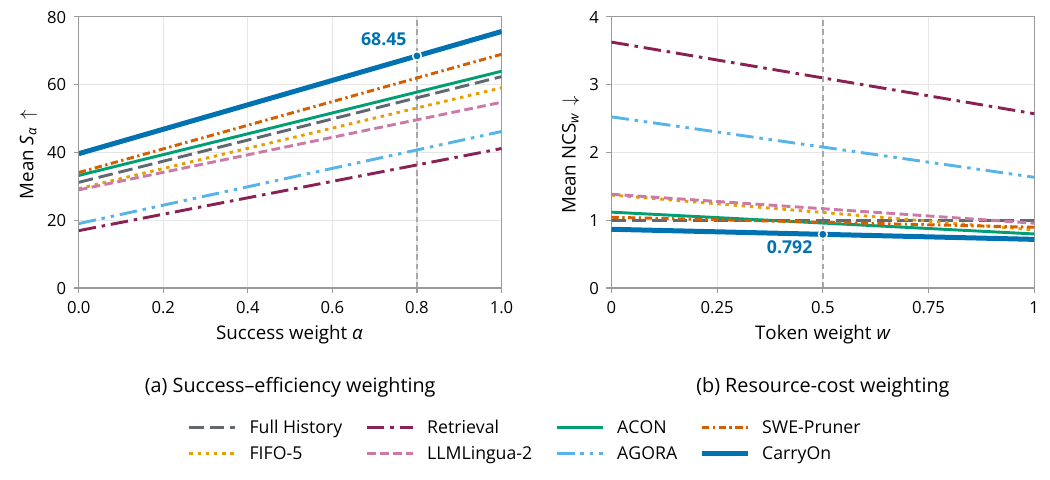}
    \caption{Sensitivity to metric weights. Scores are averaged equally
    across benchmarks and Actors. Dashed lines mark the default weights.}
    \label{fig:weight_sensitivity}
\end{figure*}

To assess whether the overall ranking depends on metric weights, we
vary each weight from 0 to 1. Figure~\ref{fig:weight_sensitivity}(a)
varies the success weight $\alpha$ in $S_\alpha$, keeping the token and
call weights within $S_0$ fixed. Panel~(b) varies the token weight $w$
in NCS, assigning weight $1-w$ to calls in Eq.~\ref{eq:ncs}.
Task outcomes and resource totals remain fixed throughout.
CarryOn retains the highest mean $S_\alpha$ and lowest mean
$\mathrm{NCS}_w$ across both ranges, showing that its overall lead
is stable under these weight changes.

\begin{table*}[tp]
\centering
\small
\setlength{\tabcolsep}{4pt}
\renewcommand{\arraystretch}{1.12}
\begin{tabular}{lrrrrrr}
\toprule
Method & TGC $\uparrow$ & Requirements & Calls & Input (M) & Output (M) & Total (M) \\
\midrule
Full History & 73/168 & 774/1,077 & 2,727 & 27.253 & 0.314 & 27.567 \\
FIFO-5 & 72/168 & 498/1,077 & 5,369 & 31.839 & 0.456 & 32.296 \\
Retrieval & 20/168 & 232/1,077 & 7,284 & 47.792 & 0.569 & 55.497 \\
LLMLingua-2 & 36/168 & 499/1,077 & 3,572 & 23.199 & 0.363 & 23.562 \\
ACON & 118/168 & 950/1,077 & 3,608 & 22.485 & 0.369 & 24.857 \\
AGORA & 93/168 & 723/1,077 & 6,012 & 35.196 & 0.684 & 35.881 \\
SWE-Pruner & 133/168 & 993/1,077 & 3,148 & 33.318 & 0.386 & 33.705 \\
CarryOn & \textbf{134/168} & 962/1,077 & 3,098 & 24.114 & 0.792 & 24.906 \\
\bottomrule
\end{tabular}
\caption{Detailed AppWorld results with Claude Sonnet 5.}
\label{tab:appworld_claude}
\end{table*}

\begin{table*}[tp]
\centering
\small
\setlength{\tabcolsep}{4pt}
\renewcommand{\arraystretch}{1.12}
\begin{tabular}{lrrrrrr}
\toprule
Method & TGC $\uparrow$ & Requirements & Calls & Input (M) & Output (M) & Total (M) \\
\midrule
Full History & 110/168 & 790/1,077 & 1,361 & 9.386 & 3.295 & 12.681 \\
FIFO-5 & 90/168 & 669/1,077 & 1,674 & 9.674 & 3.997 & 13.672 \\
Retrieval & 74/168 & 577/1,077 & 2,279 & 14.017 & 4.830 & 21.854 \\
LLMLingua-2 & 75/168 & 574/1,077 & 1,630 & 8.491 & 4.996 & 13.487 \\
ACON & 89/168 & 666/1,077 & 1,797 & 8.818 & 4.737 & 15.127 \\
AGORA & 69/168 & 552/1,077 & 1,664 & 7.867 & 6.244 & 14.111 \\
SWE-Pruner & 115/168 & 844/1,077 & 2,839 & 19.201 & 6.589 & 25.791 \\
CarryOn & \textbf{142/168} & 914/1,077 & 3,255 & 18.864 & 4.305 & 23.169 \\
\bottomrule
\end{tabular}
\caption{Detailed AppWorld results with Qwen3.8-27B.}
\label{tab:appworld_qwen}
\end{table*}

\begin{table*}[tp]
\centering
\small
\setlength{\tabcolsep}{4pt}
\renewcommand{\arraystretch}{1.12}
\begin{tabular}{lrrrrrr}
\toprule
Method & TGC $\uparrow$ & Requirements & Calls & Input (M) & Output (M) & Total (M) \\
\midrule
Full History & 90/168 & 767/1,077 & 2,401 & 16.740 & 1.386 & 18.126 \\
FIFO-5 & 82/168 & 735/1,077 & 2,820 & 13.088 & 1.434 & 14.522 \\
Retrieval & 48/168 & 454/1,077 & 5,231 & 24.511 & 1.858 & 30.800 \\
LLMLingua-2 & 70/168 & 692/1,077 & 2,428 & 11.984 & 1.459 & 13.443 \\
ACON & 82/168 & 772/1,077 & 2,650 & 12.721 & 1.311 & 15.109 \\
AGORA & 66/168 & 666/1,077 & 4,103 & 17.054 & 2.459 & 19.513 \\
SWE-Pruner & 116/168 & 918/1,077 & 2,467 & 16.213 & 1.080 & 17.293 \\
CarryOn & \textbf{121/168} & 906/1,077 & 2,983 & 19.010 & 1.623 & 20.633 \\
\bottomrule
\end{tabular}
\caption{Detailed AppWorld results with Gemma-4-31B.}
\label{tab:appworld_gemma}
\end{table*}

\begin{table*}[tp]
\centering
\small
\setlength{\tabcolsep}{4pt}
\renewcommand{\arraystretch}{1.12}
\begin{tabular}{lrrrrrr}
\toprule
Method & TGC $\uparrow$ & Conditions & Calls & Input (M) & Output (M) & Total (M) \\
\midrule
Full History & 60/95 & 113/148 & 1,484 & 29.454 & 0.410 & 29.864 \\
FIFO-5 & 55/95 & 101/141 & 1,744 & 5.680 & 0.336 & 6.016 \\
Retrieval & 47/95 & 69/117 & 2,384 & 6.793 & 0.313 & 8.483 \\
LLMLingua-2 & 61/95 & 124/158 & 1,569 & 6.067 & 0.286 & 6.353 \\
ACON & 64/95 & 131/162 & 1,574 & 7.385 & 0.272 & 8.410 \\
AGORA & 47/95 & 93/141 & 2,381 & 10.735 & 0.461 & 11.196 \\
SWE-Pruner & 59/95 & 117/153 & 1,501 & 10.733 & 0.373 & 11.106 \\
CarryOn & \textbf{66/95} & 136/165 & 1,284 & 8.643 & 0.340 & 8.983 \\
\bottomrule
\end{tabular}
\caption{Detailed OfficeBench results with Claude Sonnet 5.}
\label{tab:officebench}
\end{table*}

\begin{table*}[tp]
\centering
\small
\setlength{\tabcolsep}{4pt}
\renewcommand{\arraystretch}{1.12}
\begin{tabular}{lrrrrrr}
\toprule
Method & TGC $\uparrow$ & Conditions & Calls & Input (M) & Output (M) & Total (M) \\
\midrule
Full History & \textbf{63/95} & 128/160 & 1,531 & 10.807 & 1.668 & 12.475 \\
FIFO-5 & 51/95 & 93/137 & 1,970 & 6.638 & 1.375 & 8.013 \\
Retrieval & 49/95 & 85/131 & 2,230 & 8.411 & 1.422 & 12.729 \\
LLMLingua-2 & 57/95 & 104/142 & 1,716 & 5.729 & 1.554 & 7.282 \\
ACON & 60/95 & 116/151 & 1,611 & 4.387 & 1.856 & 6.845 \\
AGORA & 44/95 & 83/134 & 2,221 & 9.631 & 1.589 & 11.220 \\
SWE-Pruner & 60/95 & 125/160 & 1,585 & 11.834 & 2.067 & 13.901 \\
CarryOn & \textbf{63/95} & 129/161 & 935 & 3.504 & 0.742 & 4.246 \\
\bottomrule
\end{tabular}
\caption{Detailed OfficeBench results with Qwen3.8-27B.}
\label{tab:officebench_qwen}
\end{table*}

\begin{table*}[tp]
\centering
\small
\setlength{\tabcolsep}{4pt}
\renewcommand{\arraystretch}{1.12}
\begin{tabular}{lrrrrrr}
\toprule
Method & TGC $\uparrow$ & Conditions & Calls & Input (M) & Output (M) & Total (M) \\
\midrule
Full History & 49/95 & 93/139 & 1,518 & 29.123 & 0.977 & 30.100 \\
FIFO-5 & 52/95 & 105/148 & 2,110 & 3.442 & 1.242 & 4.684 \\
Retrieval & 33/95 & 60/122 & 2,938 & 4.264 & 1.285 & 6.338 \\
LLMLingua-2 & 51/95 & 118/162 & 1,575 & 3.849 & 1.067 & 4.915 \\
ACON & 55/95 & 111/151 & 1,669 & 3.926 & 0.950 & 5.114 \\
AGORA & 37/95 & 69/127 & 2,986 & 5.408 & 1.447 & 6.855 \\
SWE-Pruner & 58/95 & 129/166 & 1,690 & 6.004 & 1.031 & 7.034 \\
CarryOn & \textbf{63/95} & 139/171 & 1,078 & 4.137 & 0.659 & 4.796 \\
\bottomrule
\end{tabular}
\caption{Detailed OfficeBench results with Gemma-4-31B.}
\label{tab:officebench_gemma}
\end{table*}

\begin{table*}[tp]
\centering
\small
\setlength{\tabcolsep}{4pt}
\renewcommand{\arraystretch}{1.12}
\begin{tabular}{lrrrrrr}
\toprule
Method & Success & \shortstack{Pass$^1$ /\\Avg. Reward} & \shortstack{Calls\\A/U/C} & Input (M) & Output (M) & Total (M) \\
\midrule
Full History & 30/40 & 0.750 & 665/853/0 & 13.287 & 0.136 & 13.423 \\
FIFO-5 & 29/40 & 0.725 & 816/963/0 & 15.385 & 0.171 & 15.556 \\
Retrieval & 17/40 & 0.425 & 1,621/1,301/0 & 28.615 & 0.276 & 28.891 \\
LLMLingua-2 & \textbf{32/40} & 0.800 & 694/878/7 & 13.977 & 0.145 & 14.122 \\
ACON & 29/40 & 0.725 & 641/799/2 & 12.647 & 0.135 & 12.782 \\
AGORA & 25/40 & 0.625 & 1,087/1,075/0 & 19.423 & 0.234 & 19.657 \\
SWE-Pruner & 28/40 & 0.700 & 638/820/0 & 13.363 & 0.136 & 13.499 \\
CarryOn & 30/40 & 0.750 & 676/825/0 & 16.175 & 0.276 & 16.451 \\
\bottomrule
\end{tabular}
\caption{Detailed Tau2 Telecom results with Claude Sonnet 5. A/U/C denote Actor, User, and compressor calls.}
\label{tab:tau2}
\end{table*}

\begin{table*}[tp]
\centering
\small
\setlength{\tabcolsep}{4pt}
\renewcommand{\arraystretch}{1.12}
\begin{tabular}{lrrrrrr}
\toprule
Method & Success & Avg. Reward & \shortstack{Calls\\A/U/C} & Input (M) & Output (M) & Total (M) \\
\midrule
Full History & 32/40 & 0.800 & 762/890/0 & 12.619 & 0.305 & 12.924 \\
FIFO-5 & 35/40 & 0.875 & 818/981/0 & 13.385 & 0.340 & 13.725 \\
Retrieval & 29/40 & 0.725 & 995/975/0 & 14.615 & 0.386 & 15.002 \\
LLMLingua-2 & 32/40 & 0.800 & 778/885/5 & 12.361 & 0.260 & 12.621 \\
ACON & 35/40 & 0.875 & 730/859/3 & 11.607 & 0.276 & 11.883 \\
AGORA & 20/40 & 0.500 & 2,042/1,623/0 & 27.498 & 0.805 & 28.303 \\
SWE-Pruner & 33/40 & 0.825 & 800/871/0 & 13.289 & 0.289 & 13.578 \\
CarryOn & \textbf{36/40} & 0.900 & 600/761/0 & 12.746 & 0.415 & 13.161 \\
\bottomrule
\end{tabular}
\caption{Detailed Tau2 Telecom results with Qwen3.8-27B. A/U/C denote Actor, User, and compressor calls.}
\label{tab:tau2_qwen}
\end{table*}

\begin{table*}[tp]
\centering
\small
\setlength{\tabcolsep}{4pt}
\renewcommand{\arraystretch}{1.12}
\begin{tabular}{lrrrrrr}
\toprule
Method & Success & \shortstack{Pass$^1$ /\\Avg. Reward} & \shortstack{Calls\\A/U/C} & Input (M) & Output (M) & Total (M) \\
\midrule
Full History & 25/40 & 0.625 & 561/647/0 & 8.658 & 0.224 & 8.882 \\
FIFO-5 & 24/40 & 0.600 & 601/649/0 & 8.231 & 0.256 & 8.487 \\
Retrieval & 14/40 & 0.350 & 1,187/657/0 & 14.266 & 0.513 & 14.779 \\
LLMLingua-2 & 19/40 & 0.475 & 622/659/5 & 9.075 & 0.239 & 9.314 \\
ACON & 22/40 & 0.550 & 719/765/5 & 10.747 & 0.263 & 11.010 \\
AGORA & 13/40 & 0.325 & 1,087/960/0 & 13.921 & 0.430 & 14.351 \\
SWE-Pruner & 26/40 & 0.650 & 504/608/0 & 7.775 & 0.215 & 7.990 \\
CarryOn & \textbf{31/40} & 0.775 & 510/633/0 & 8.755 & 0.393 & 9.148 \\
\bottomrule
\end{tabular}
\caption{Detailed Tau2 Telecom results with Gemma-4-31B. A/U/C denote Actor, User, and compressor calls.}
\label{tab:tau2_gemma}
\end{table*}

\section{Context and Execution Cost Across Actors}
\label{sec:cross_dataset_context}
\label{sec:context_cost_curves}

\begin{figure*}[!tp]
\centering
\includegraphics[width=\textwidth]{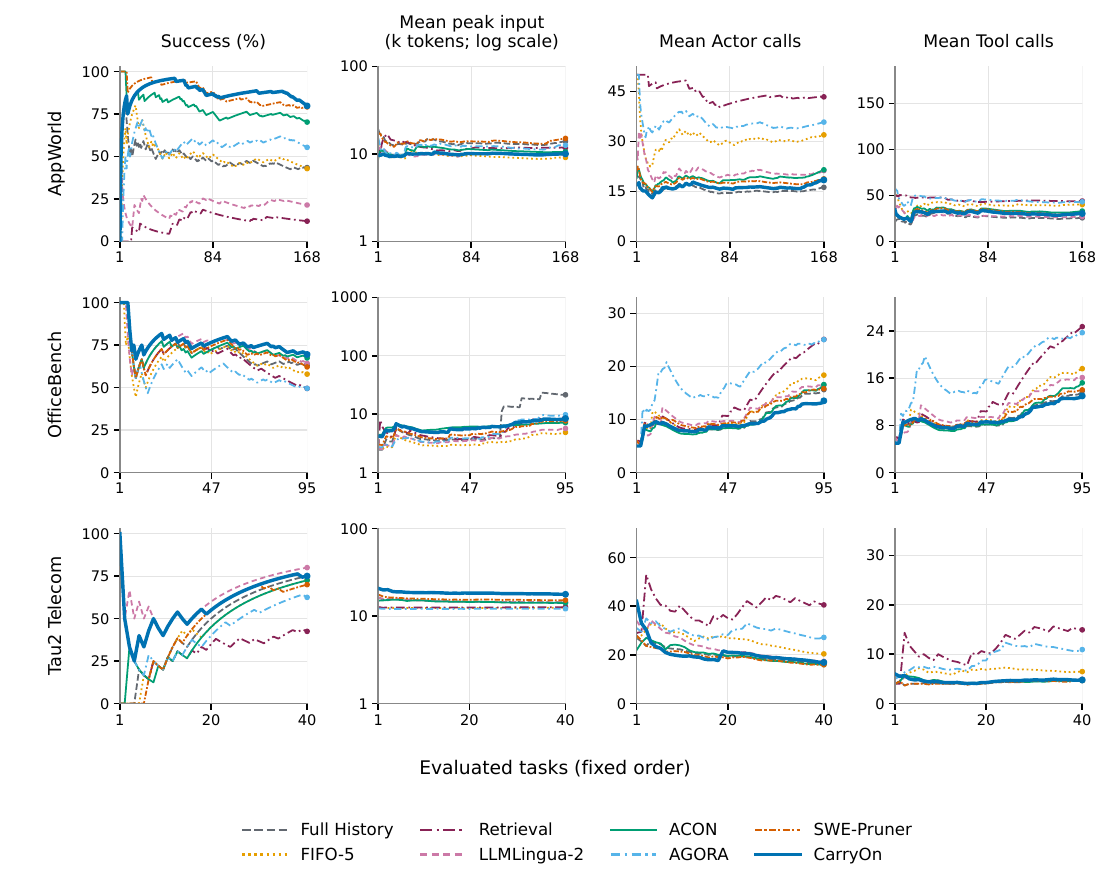}
\caption{Claude Sonnet 5: Success and resource use across three benchmarks,
averaged over fixed-order task prefixes.}
\label{fig:claude_context_calls}
\end{figure*}

\begin{figure*}[!tp]
\centering
\includegraphics[width=\textwidth]{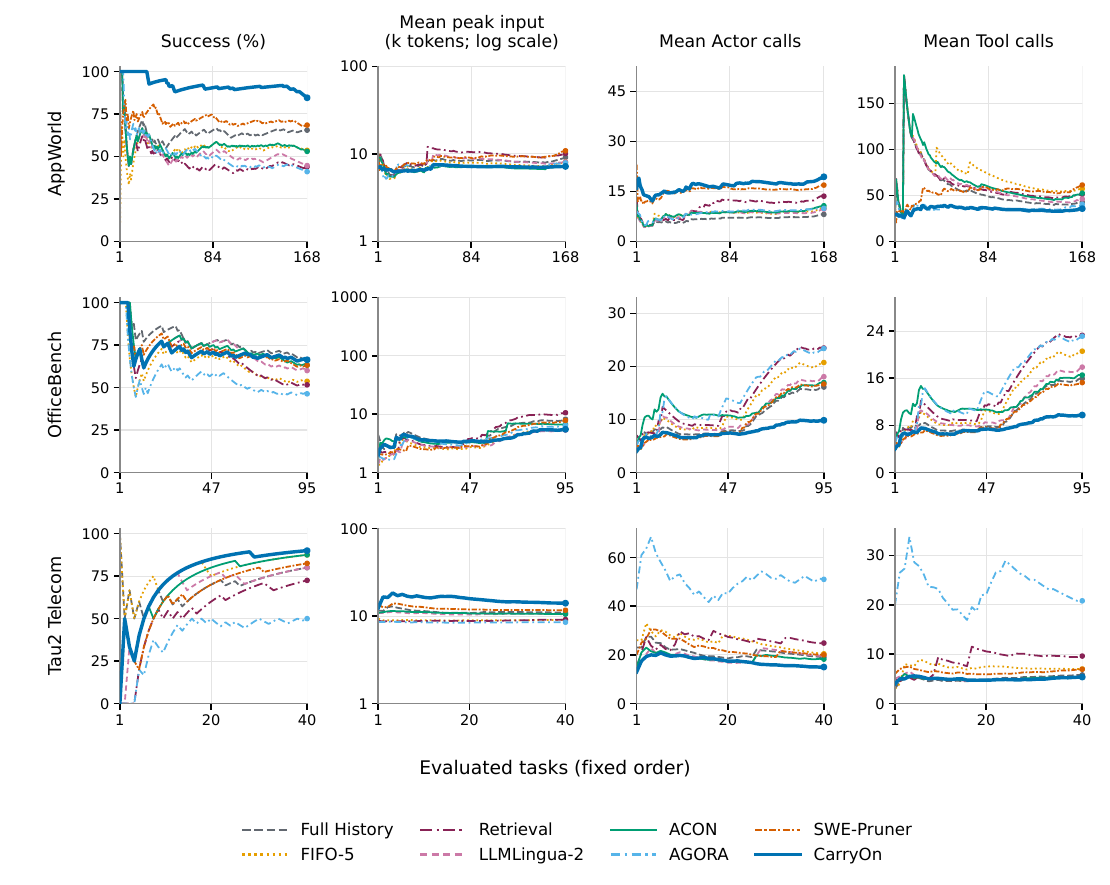}
\caption{Qwen3.8-27B: Success and resource use across three benchmarks,
with the same metrics and task order as Figure~\ref{fig:claude_context_calls}.}
\label{fig:qwen_context_calls}
\end{figure*}

\begin{figure*}[!tp]
\centering
\includegraphics[width=\textwidth]{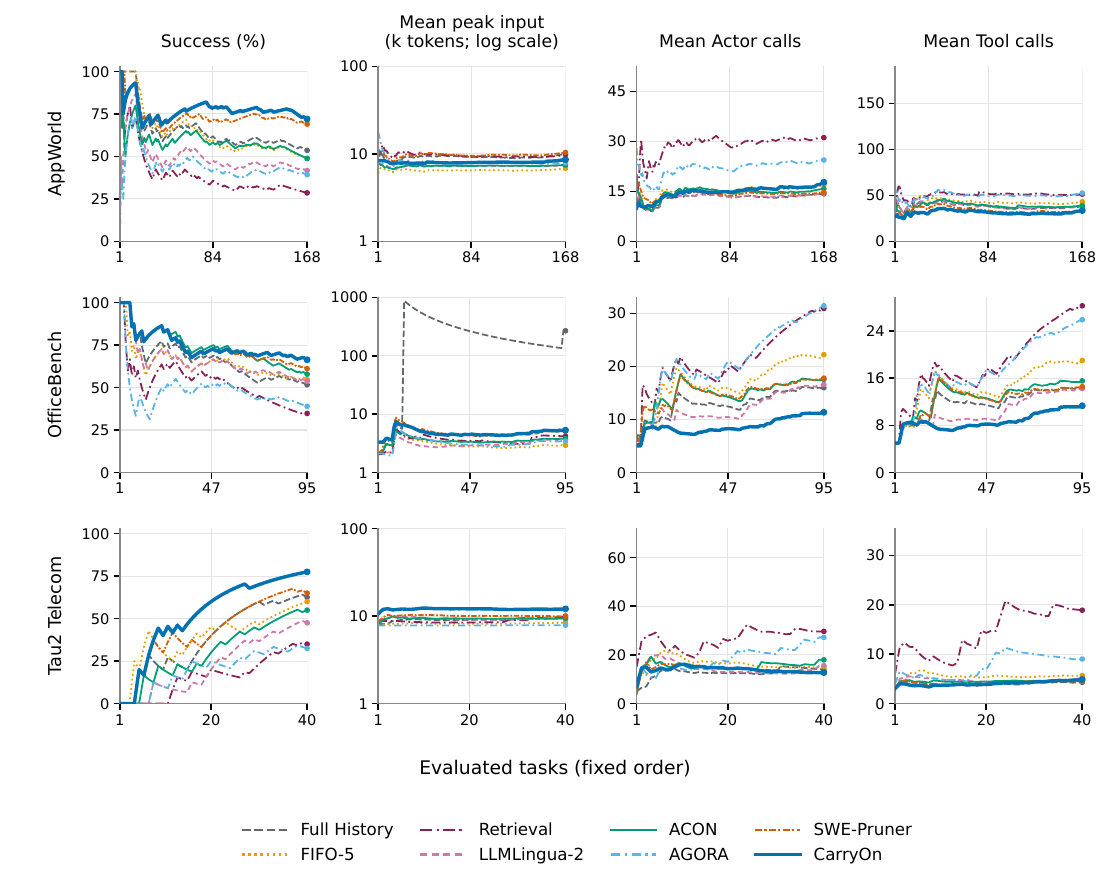}
\caption{Gemma-4-31B: Success and resource use across three benchmarks,
with the same metrics and task order as Figure~\ref{fig:claude_context_calls}.}
\label{fig:gemma_context_calls}
\end{figure*}

Figures~\ref{fig:claude_context_calls}--\ref{fig:gemma_context_calls}
compare all eight methods using the runs in Table~\ref{tab:main_results}.
Each point averages the first $n$ tasks in a fixed benchmark order,
including failures, and the final point gives the full-set mean.
The horizontal axis shows the number of evaluated tasks.

A task's peak input is its largest submitted Actor input, including
fixed prompts. Task-level peaks are averaged arithmetically and plotted
on logarithmic axes, with matching scales across Actors. Tool calls follow
Appendix~\ref{sec:cost_accounting_details}.

Over the complete OfficeBench task set, CarryOn matches or exceeds the
best baseline Success while using the fewest Actor and tool calls per
task for each Actor. On AppWorld, it improves Success and lowers mean
peak input relative to Full History for all three Actors. On Tau2 with
Qwen, it increases Success from 80\% to 90\% while reducing mean Actor
calls from 19.05 to 15.00 relative to Full History.

On Tau2, CarryOn has higher mean peak input than Full History.

\twocolumn[{%
\begin{minipage}{\textwidth}
    \captionsetup{hypcap=false,type=figure}
    \centering
    \includegraphics[width=0.85\textwidth]{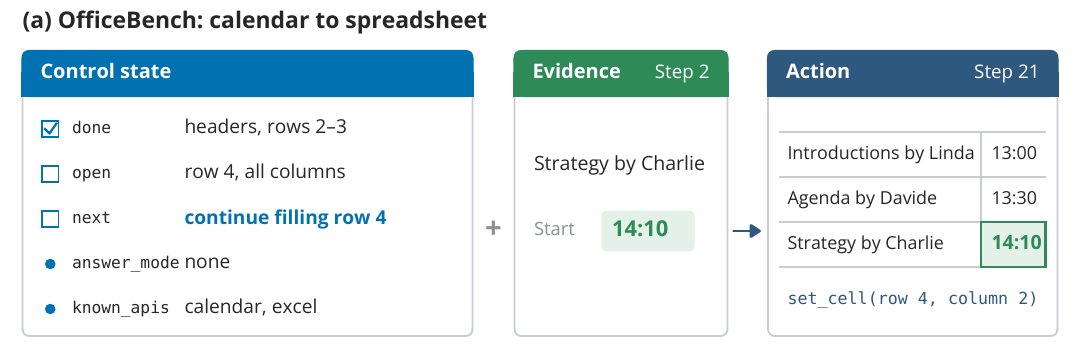}
    \par\smallskip
    \includegraphics[width=0.85\textwidth]{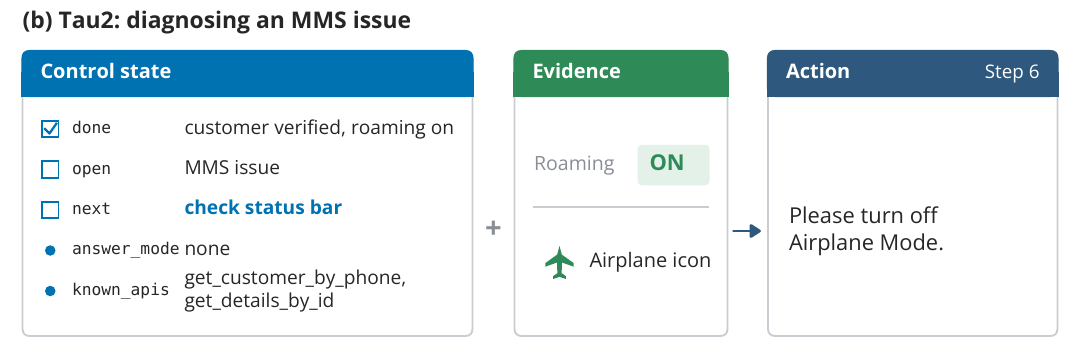}
    \caption{Control state and evidence across environments.
    (a) An earlier calendar record supplies a spreadsheet value.
    (b) Account information and a new user report support the next diagnostic reply.
    Text and timestamps are abbreviated.}
    \label{fig:cross_environment_examples}
\end{minipage}
\vspace{0.6em}
}]

\section{Control State and Evidence Across Environments}
\label{sec:cross_environment_examples}

The following examples from recorded Claude Sonnet 5 trajectories show
how control state and evidence support actions in OfficeBench and Tau2
(Figure~\ref{fig:cross_environment_examples}).

\subsection{OfficeBench: Transferring a Calendar Agenda to a Spreadsheet}
In Figure~\ref{fig:cross_environment_examples}(a), the Actor is transferring
a calendar agenda to a spreadsheet. The control state records completed
headers and rows 2--3, with row 4 still in progress. The evidence pack
retains the calendar record obtained at step 2, which gives 14:10 as the
start time of ``Strategy by Charlie.'' At step 21, the Actor uses this
value to fill row 4, column 2.

\subsection{Tau2: Continuing a Customer-Service Diagnosis}
In Figure~\ref{fig:cross_environment_examples}(b), a customer cannot send
MMS messages while abroad. The control state records that the customer
has been verified, roaming is enabled, and the MMS issue remains open.
It identifies checking the status bar as the next diagnostic step. The
available evidence includes the account record and the customer's new
report of an airplane icon. This report lets the Actor advance the
diagnosis: at step 6, it asks the customer to turn off Airplane Mode.

\section{Evidence Reuse and Case Studies}
\label{sec:evidence_cases}

\subsection{Long-Term Evidence Retention and Reselection}
\label{sec:evidence_reuse_details}

\paragraph{Coverage and counting.}
We use the designated CarryOn runs reported in the main table, including
successful and unsuccessful trajectories. All 909 trajectories and
14,419 Actor calls have complete selection records. Each occurrence of
a selected unit at an Actor decision is counted once. Figure~\ref{fig:evidence_reuse}
pools these occurrences within each benchmark. Age is the number of
Actor steps since the source interaction, so the latest observation
has age one. For Tau2, we map a unit to the first Actor boundary at
which it enters the archive.
The distribution includes older evidence selected by the learned
manager, including temporary expansion, and excludes automatically
retained recent evidence. On Tau2, these age statistics cover the 13 of
120 trajectories that invoke learned selection. The remaining trajectories
retain the full history within the base budget.

\paragraph{Evidence reselection.}
We count a reselection when previously visible evidence is omitted from
the evidence pack for at least one decision and later selected again.

\paragraph{Retention and reselection statistics.}
Table~\ref{tab:evidence_reuse_statistics} reports the share of selected
evidence originating at least ten Actor steps earlier and the number of
task runs in which evidence is selected again after leaving the pack.
Such reselection occurs in 365 of 909 trajectories (40.2\%), showing
that evidence can return to the working context after being omitted.
The following payment example shows how the Actor uses a reselected
card record to add a payment card and refill the balance.

\begin{table*}[t]
    \centering
    \small
    \setlength{\tabcolsep}{5pt}
    \begin{tabular}{llrrr}
        \toprule
        Benchmark & Actor & Task runs & \shortstack{Selected evidence\\$\geq 10$ steps old (\%)} & \shortstack{Runs with evidence\\reselection} \\
        \midrule
        AppWorld & Claude & 168 & 31.4 & 96/168 \\
        AppWorld & Qwen & 168 & 32.5 & 94/168 \\
        AppWorld & Gemma & 168 & 27.6 & 105/168 \\
        \midrule
        OfficeBench & Claude & 95 & 40.5 & 22/95 \\
        OfficeBench & Qwen & 95 & 19.2 & 22/95 \\
        OfficeBench & Gemma & 95 & 28.0 & 13/95 \\
        \midrule
        Tau2 Telecom & Claude & 40 & 81.7 & 2/40 \\
        Tau2 Telecom & Qwen & 40 & 28.4 & 11/40 \\
        Tau2 Telecom & Gemma & 40 & -- & 0/40 \\
        \bottomrule
    \end{tabular}
    \caption{Long-term evidence retention and reselection.
    Evidence age is measured in Actor steps, with repeated selections counted
    separately. The last column reports runs with evidence selected again
    after leaving the pack, out of all runs, including failures.
    ``--'' indicates no learned-selection events.}
    \label{tab:evidence_reuse_statistics}
\end{table*}

\begin{figure}[t]
    \centering
    \includegraphics[width=\columnwidth]{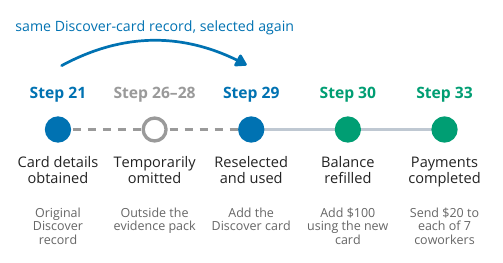}
    \caption{A payment-card record returns to the evidence pack and is
    used to add a card, refill the balance, and complete the payments.}
    \label{fig:evidence_reuse_case}
\end{figure}

\paragraph{Payment case.}
Figure~\ref{fig:evidence_reuse_case} traces a payment-card record from
its initial observation to its later selection and use.
For task \texttt{f861c32\_2}, the Discover record
(\texttt{E021/\$/2}) is obtained at step~21 and visible at steps~22--25.
It is absent at steps~26--28 and returns through learned ranking at
step~29, outside both recent retention and temporary expansion.
The saved action at step~29 selects the Discover entry from the
previously stored card list and passes its six fields to
\texttt{add\_payment\_card}. The tool returns card ID~479, which the next
action uses to add \$100 to the balance. Seven successful payment
receipts follow at step~33, and the trajectory passes the official
completion check.

\subsection{Comparative Case Studies}
\label{sec:comparative_evidence_cases}
We examine three AppWorld tasks, comparing CarryOn with ACON using
Claude Sonnet 5 in Cases~1 and~3, and with SWE-Pruner using Gemma-4-31B
in Case~2.

\begin{figure*}[t]
    \centering
    \includegraphics[width=0.85\textwidth]{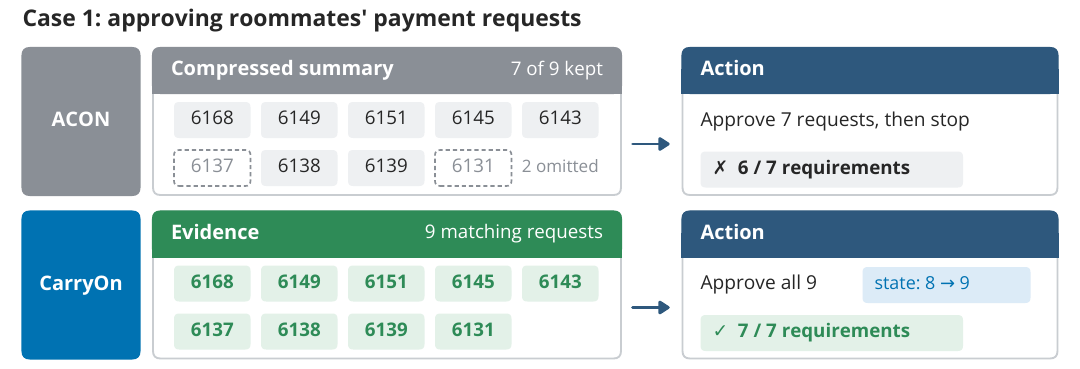}
    \caption{Case~1 (AppWorld, Claude Sonnet 5). ACON's summary omits two
    qualifying requests, shown with dashed outlines, while CarryOn retains
    the complete list.}
    \label{fig:case_payment_requests}
\end{figure*}

\paragraph{Case 1: preserving a complete payment-request list.}
The task is to approve all payment requests from roommates in the current
month. Nine of the 47 pending requests qualify
(Figure~\ref{fig:case_payment_requests}). ACON's compressed summary omits
IDs 6137 and 6131, and the Actor approves only the remaining seven
requests, passing six of seven requirements. CarryOn retains the filtered
nine-ID list as evidence and corrects the expected request count in its
state from eight to nine. It ultimately approves all nine requests and
passes all seven requirements.

\begin{figure*}[!t]
    \centering
    \includegraphics[width=0.85\textwidth]{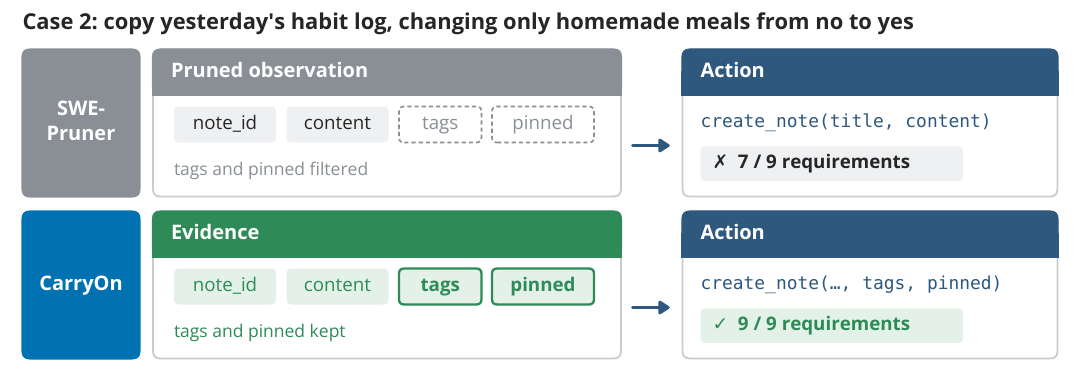}
    \caption{Case~2 (AppWorld, Gemma-4-31B). Dashed outlines mark fields
    filtered by SWE-Pruner and preserved by CarryOn. Fractions report
    passed task requirements.}
    \label{fig:case_habit_log}
\end{figure*}

\paragraph{Case 2: preserving note details.}
The task is to copy yesterday's habit log into a new note for today,
changing only the homemade-meals entry from no to yes
(Figure~\ref{fig:case_habit_log}). SWE-Pruner filters the tags and pinned
status from the returned note. The Actor updates the meal entry correctly
but omits these two fields when creating the new note, passing seven of
nine requirements. CarryOn retains the original note with both fields
and copies them alongside the updated content, passing all nine requirements.

\paragraph{Case 3: updating an expense note.}
The task is to add the current month's expenses to a note whose entries
are ordered newest first. CarryOn keeps the original note as evidence,
inserts the new month directly below the header, and passes all ten
requirements. ACON also updates the note correctly but fails the final
answer check, passing nine of ten.

\subsection{Temporary Evidence Expansion}
\label{sec:expansion_case}
In this AppWorld example with Gemma, the Actor must text a favorite
recipe's name to the user's husband. The earlier contact record
does not fit within the remaining base budget, but the selector
predicts that it is needed for the next action. Temporary evidence
expansion therefore adds it beyond the base budget
(Section~\ref{sec:learned_selection}).

\par\medskip
\noindent\begin{minipage}{\columnwidth}
\small
\begin{verbatim}
[26] Observe husband's contact
     phone_number = <number>

[31] Base budget: nearly full
     [=================== ]
     Base pack: contact omitted
     Next-action use: 0.695 > 0.5
     Expansion: add contact record
     Action: send recipe name to <number>
     Feedback: "Text message sent."

Outcome: task completed successfully
\end{verbatim}
\end{minipage}
\par\medskip
The expansion restores the contact record for the current decision,
and the Actor uses its phone number in the same step.

\section{Performance Across Task Groups}
\label{sec:stratified_results}
\begin{figure*}[!tp]
    \centering
    \includegraphics[width=\textwidth]{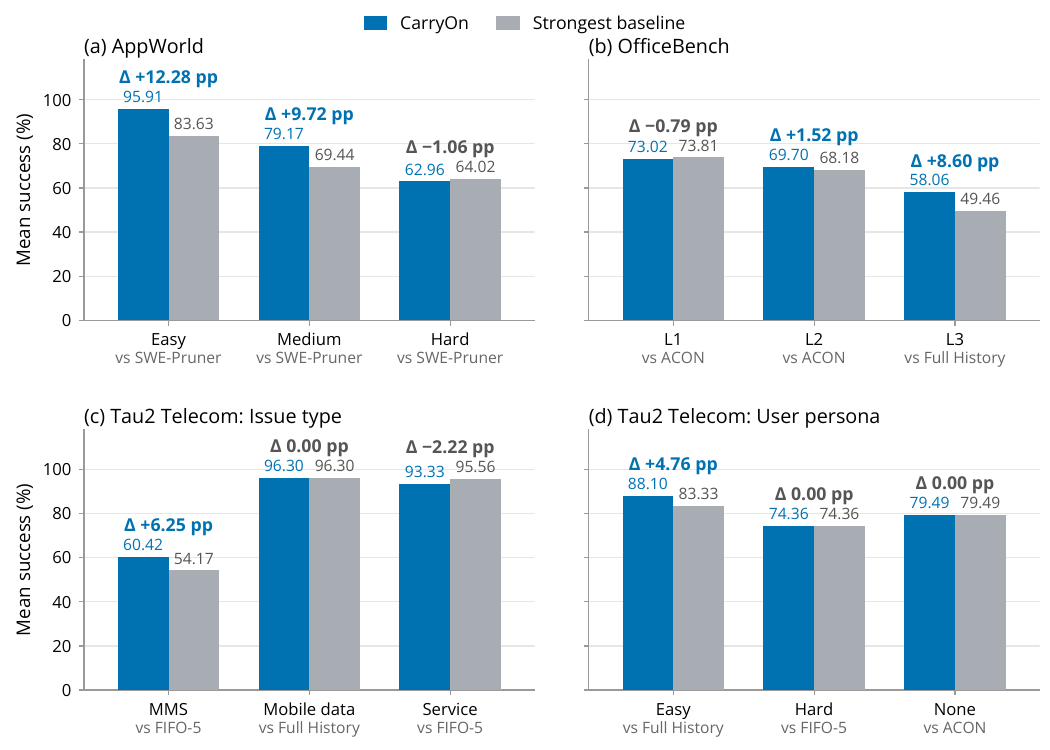}
    \caption{Success by task group, averaged across Actors. The strongest
    baseline is selected separately for each group. $\Delta$ denotes the
    difference in percentage points.}
    \label{fig:task_groups}
\end{figure*}

Figure~\ref{fig:task_groups} summarizes Success across task groups.
Tables~\ref{tab:appworld_difficulty}--\ref{tab:tau2_strata} report all
methods and Actors, retaining each dataset's original denominators.
AppWorld difficulty, OfficeBench level, and Tau2 persona describe
different task properties. Tau2 issue type and persona are two
partitions of the same 40 tasks.

\begin{table*}[tp]
\centering
\small
\setlength{\tabcolsep}{4pt}
\renewcommand{\arraystretch}{1.12}
\begin{tabular}{llrrrr}
\toprule
Actor & Method & Easy & Medium & Hard & Overall \\
\midrule
Claude Sonnet 5 & Full History & 34/57 & 21/48 & 18/63 & 73/168 \\
Claude Sonnet 5 & FIFO-5 & 49/57 & 17/48 & 6/63 & 72/168 \\
Claude Sonnet 5 & Retrieval & 16/57 & 1/48 & 3/63 & 20/168 \\
Claude Sonnet 5 & LLMLingua-2 & 24/57 & 7/48 & 5/63 & 36/168 \\
Claude Sonnet 5 & ACON & 48/57 & 36/48 & 34/63 & 118/168 \\
Claude Sonnet 5 & AGORA & 48/57 & 23/48 & 22/63 & 93/168 \\
Claude Sonnet 5 & SWE-Pruner & 47/57 & 40/48 & 46/63 & 133/168 \\
Claude Sonnet 5 & CarryOn & 52/57 & 41/48 & 41/63 & 134/168 \\
\midrule
Qwen3.8-27B & Full History & 38/57 & 30/48 & 42/63 & 110/168 \\
Qwen3.8-27B & FIFO-5 & 37/57 & 29/48 & 24/63 & 90/168 \\
Qwen3.8-27B & Retrieval & 33/57 & 24/48 & 17/63 & 74/168 \\
Qwen3.8-27B & LLMLingua-2 & 37/57 & 22/48 & 16/63 & 75/168 \\
Qwen3.8-27B & ACON & 37/57 & 28/48 & 24/63 & 89/168 \\
Qwen3.8-27B & AGORA & 31/57 & 21/48 & 17/63 & 69/168 \\
Qwen3.8-27B & SWE-Pruner & 47/57 & 32/48 & 36/63 & 115/168 \\
Qwen3.8-27B & CarryOn & 57/57 & 46/48 & 39/63 & 142/168 \\
\midrule
Gemma-4-31B & Full History & 47/57 & 23/48 & 20/63 & 90/168 \\
Gemma-4-31B & FIFO-5 & 44/57 & 21/48 & 17/63 & 82/168 \\
Gemma-4-31B & Retrieval & 30/57 & 15/48 & 3/63 & 48/168 \\
Gemma-4-31B & LLMLingua-2 & 44/57 & 15/48 & 11/63 & 70/168 \\
Gemma-4-31B & ACON & 42/57 & 22/48 & 18/63 & 82/168 \\
Gemma-4-31B & AGORA & 43/57 & 12/48 & 11/63 & 66/168 \\
Gemma-4-31B & SWE-Pruner & 49/57 & 28/48 & 39/63 & 116/168 \\
Gemma-4-31B & CarryOn & 55/57 & 27/48 & 39/63 & 121/168 \\
\bottomrule
\end{tabular}
\caption{Successful tasks by AppWorld difficulty.}
\label{tab:appworld_difficulty}
\end{table*}

\begin{table*}[tp]
\centering
\small
\setlength{\tabcolsep}{4pt}
\renewcommand{\arraystretch}{1.12}
\begin{tabular}{llrrrrr}
\toprule
Actor & Method & L1 TGC & L2 TGC & L3 TGC & Passed/checked & Unchecked \\
\midrule
Claude Sonnet 5 & Full History & 31/42 & 12/22 & 17/31 & 113/148 & 61 \\
Claude Sonnet 5 & FIFO-5 & 31/42 & 13/22 & 11/31 & 101/141 & 68 \\
Claude Sonnet 5 & Retrieval & 32/42 & 10/22 & 5/31 & 69/117 & 92 \\
Claude Sonnet 5 & LLMLingua-2 & 33/42 & 14/22 & 14/31 & 124/158 & 51 \\
Claude Sonnet 5 & ACON & 30/42 & 17/22 & 17/31 & 131/162 & 47 \\
Claude Sonnet 5 & AGORA & 26/42 & 10/22 & 11/31 & 93/141 & 68 \\
Claude Sonnet 5 & SWE-Pruner & 31/42 & 14/22 & 14/31 & 117/153 & 56 \\
Claude Sonnet 5 & CarryOn & 32/42 & 16/22 & 18/31 & 136/165 & 44 \\
\midrule
Qwen3.8-27B & Full History & 32/42 & 14/22 & 17/31 & 128/160 & 49 \\
Qwen3.8-27B & FIFO-5 & 29/42 & 10/22 & 12/31 & 93/137 & 72 \\
Qwen3.8-27B & Retrieval & 32/42 & 8/22 & 9/31 & 85/131 & 78 \\
Qwen3.8-27B & LLMLingua-2 & 32/42 & 14/22 & 11/31 & 104/142 & 67 \\
Qwen3.8-27B & ACON & 32/42 & 14/22 & 14/31 & 116/151 & 58 \\
Qwen3.8-27B & AGORA & 25/42 & 7/22 & 12/31 & 83/134 & 75 \\
Qwen3.8-27B & SWE-Pruner & 31/42 & 13/22 & 16/31 & 125/160 & 49 \\
Qwen3.8-27B & CarryOn & 30/42 & 15/22 & 18/31 & 129/161 & 48 \\
\midrule
Gemma-4-31B & Full History & 29/42 & 8/22 & 12/31 & 93/139 & 70 \\
Gemma-4-31B & FIFO-5 & 27/42 & 12/22 & 13/31 & 105/148 & 61 \\
Gemma-4-31B & Retrieval & 25/42 & 6/22 & 2/31 & 60/122 & 87 \\
Gemma-4-31B & LLMLingua-2 & 27/42 & 11/22 & 13/31 & 118/162 & 47 \\
Gemma-4-31B & ACON & 31/42 & 14/22 & 10/31 & 111/151 & 58 \\
Gemma-4-31B & AGORA & 22/42 & 8/22 & 7/31 & 69/127 & 82 \\
Gemma-4-31B & SWE-Pruner & 30/42 & 14/22 & 14/31 & 129/166 & 43 \\
Gemma-4-31B & CarryOn & 30/42 & 15/22 & 18/31 & 139/171 & 38 \\
\bottomrule
\end{tabular}
\caption{Successful tasks and evaluator coverage by OfficeBench level. Unchecked conditions were not evaluated.}
\label{tab:officebench_levels}
\end{table*}

\begin{table*}[tp]
\centering
\small
\setlength{\tabcolsep}{4pt}
\renewcommand{\arraystretch}{1.12}
\begin{tabular}{llrrrrrr}
\toprule
Actor & Method & Easy & Hard & None & MMS & Mobile data & Service \\
\midrule
Claude Sonnet 5 & Full History & 11/14 & 10/13 & 9/13 & 6/16 & 9/9 & 15/15 \\
Claude Sonnet 5 & FIFO-5 & 9/14 & 9/13 & 11/13 & 7/16 & 7/9 & 15/15 \\
Claude Sonnet 5 & Retrieval & 5/14 & 6/13 & 6/13 & 5/16 & 4/9 & 8/15 \\
Claude Sonnet 5 & LLMLingua-2 & 13/14 & 10/13 & 9/13 & 8/16 & 9/9 & 15/15 \\
Claude Sonnet 5 & ACON & 11/14 & 8/13 & 10/13 & 5/16 & 9/9 & 15/15 \\
Claude Sonnet 5 & AGORA & 9/14 & 9/13 & 7/13 & 6/16 & 6/9 & 13/15 \\
Claude Sonnet 5 & SWE-Pruner & 10/14 & 9/13 & 9/13 & 7/16 & 9/9 & 12/15 \\
Claude Sonnet 5 & CarryOn & 12/14 & 9/13 & 9/13 & 8/16 & 8/9 & 14/15 \\
\midrule
Qwen3.8-27B & Full History & 13/14 & 10/13 & 9/13 & 10/16 & 8/9 & 14/15 \\
Qwen3.8-27B & FIFO-5 & 14/14 & 11/13 & 10/13 & 13/16 & 7/9 & 15/15 \\
Qwen3.8-27B & Retrieval & 9/14 & 9/13 & 11/13 & 9/16 & 7/9 & 13/15 \\
Qwen3.8-27B & LLMLingua-2 & 13/14 & 9/13 & 10/13 & 11/16 & 8/9 & 13/15 \\
Qwen3.8-27B & ACON & 13/14 & 10/13 & 12/13 & 12/16 & 9/9 & 14/15 \\
Qwen3.8-27B & AGORA & 7/14 & 6/13 & 7/13 & 8/16 & 4/9 & 8/15 \\
Qwen3.8-27B & SWE-Pruner & 12/14 & 8/13 & 13/13 & 10/16 & 9/9 & 14/15 \\
Qwen3.8-27B & CarryOn & 13/14 & 11/13 & 12/13 & 13/16 & 9/9 & 14/15 \\
\midrule
Gemma-4-31B & Full History & 11/14 & 7/13 & 7/13 & 4/16 & 9/9 & 12/15 \\
Gemma-4-31B & FIFO-5 & 8/14 & 9/13 & 7/13 & 6/16 & 5/9 & 13/15 \\
Gemma-4-31B & Retrieval & 3/14 & 5/13 & 6/13 & 3/16 & 1/9 & 10/15 \\
Gemma-4-31B & LLMLingua-2 & 7/14 & 6/13 & 6/13 & 2/16 & 6/9 & 11/15 \\
Gemma-4-31B & ACON & 8/14 & 5/13 & 9/13 & 3/16 & 7/9 & 12/15 \\
Gemma-4-31B & AGORA & 4/14 & 4/13 & 5/13 & 4/16 & 2/9 & 7/15 \\
Gemma-4-31B & SWE-Pruner & 9/14 & 9/13 & 8/13 & 6/16 & 7/9 & 13/15 \\
Gemma-4-31B & CarryOn & 12/14 & 9/13 & 10/13 & 8/16 & 9/9 & 14/15 \\
\bottomrule
\end{tabular}
\caption{Successful Tau2 tasks by user persona and issue type.}
\label{tab:tau2_strata}
\end{table*}

\section{CarryOn-E}
\label{sec:local_distillation}
\begin{table*}[tp]
\centering
\small
\setlength{\tabcolsep}{4pt}
\renewcommand{\arraystretch}{1.15}
\begin{tabular*}{\textwidth}{@{\extracolsep{\fill}}lrrrr@{}}
\toprule
\multicolumn{5}{l}{\textbf{Overall results}} \\
Dataset & Success (\%) & Completed/total & Actor calls & Tokens (M) \\
\midrule
AppWorld & 47.0 & 79/168 & 4,961 & 30.982 \\
OfficeBench & 46.3 & 44/95 & 1,219 & 7.311 \\
Tau2 Telecom & 67.5 & 27/40 & 998 & 19.473 \\
\bottomrule
\end{tabular*}
\medskip
\begin{tabular*}{\textwidth}{@{\extracolsep{\fill}}llccc@{}}
\toprule
\multicolumn{5}{l}{\textbf{Success by task group}} \\
Dataset & Grouping & \multicolumn{3}{c}{Success (\%); completed/total} \\
\midrule
AppWorld & Difficulty & \shortstack{Easy\\70.2 (40/57)} & \shortstack{Medium\\62.5 (30/48)} & \shortstack{Hard\\14.3 (9/63)} \\
\addlinespace
OfficeBench & Level & \shortstack{L1\\50.0 (21/42)} & \shortstack{L2\\50.0 (11/22)} & \shortstack{L3\\38.7 (12/31)} \\
\addlinespace
Tau2 Telecom & Issue type & \shortstack{MMS\\75.0 (12/16)} & \shortstack{Mobile data\\88.9 (8/9)} & \shortstack{Service\\46.7 (7/15)} \\
\addlinespace
Tau2 Telecom & User persona & \shortstack{Easy\\71.4 (10/14)} & \shortstack{Hard\\61.5 (8/13)} & \shortstack{None\\69.2 (9/13)} \\
\bottomrule
\end{tabular*}
\caption{CarryOn-E (9B) results across benchmarks and task groups.}
\label{tab:carryon_e_results}
\end{table*}

CarryOn-E explores local deployment by replacing the \emph{LLM Actor}
in region~3 of Figure~\ref{fig:method} with a smaller model while keeping
the selector and retention policy fixed. We distill a frozen
Qwen3.8-27B CarryOn Actor into a Qwen3.5-9B student using supervised
fine-tuning followed by DAgger-style distillation.

\subsection{Training Procedure}
Given Actor context $X$, the student policy $\pi_\theta$ learns the
teacher's visible response $v=(v_1,\ldots,v_L)$, containing the MEMORY
update and an action in the environment's required format.
We minimize the response loss
\begin{equation}
    \ell_{\mathrm{actor}}
    =-\frac{1}{L}\sum_{j=1}^{L}
    \log\pi_\theta(v_j\mid X,v_{<j}),
    \label{eq:actor_distillation_loss}
\end{equation}
where $\theta$ denotes the student's LoRA parameters, $L$ is the response
length in tokens, and $v_{<j}$ contains the preceding response tokens.
Initial supervised fine-tuning uses context--response pairs $\mathcal D_0$
from successful teacher trajectories on training tasks to obtain $\theta_0$.
The loss is computed only over response tokens.

A student can reach states that are absent from successful teacher
trajectories. To train on these states, we apply DAgger-style
distillation~\citep{ross2011reduction}. In each round, the student performs
training tasks using CarryOn to construct its context. The teacher
provides a response for each visited context, including those from
unsuccessful rollouts. These context--response pairs form $\mathcal B_k$,
which is added to the accumulated dataset to further train the student:
\begin{equation}
\begin{aligned}
    \mathcal D_k &= \mathcal D_{k-1}\cup\mathcal B_k,\\
    \theta_k &\leftarrow
    \operatorname{FineTune}(\theta_{k-1},\mathcal D_k).
\end{aligned}
\label{eq:actor_dataset_aggregation}
\end{equation}
Here, $\operatorname{FineTune}$ continues the student's LoRA adapter on
the accumulated data using Eq.~\ref{eq:actor_distillation_loss}, which
averages over target tokens. Training balances examples across tasks or
source trajectories.

\subsection{Training and Evaluation Settings}
The student is initialized from Qwen3.5-9B and evaluated on all three
datasets. Each aggregation round trains for one epoch over the accumulated
data. The reported models use six rounds on AppWorld, seven on
OfficeBench, and five on Tau2.
Student inference uses non-thinking mode, and distillation uses only
training-split tasks.
Estimated cumulative training wall times, including initial supervised
fine-tuning and all aggregation rounds leading to the reported models,
are 54.00, 23.06, and 54.28 hours on AppWorld, OfficeBench, and Tau2,
respectively, using one H100 GPU per training job. These estimates
exclude rollout collection and teacher labeling.

\subsection{Local Deployment Results}
\label{sec:local_deployment}
CarryOn-E offers a smaller-model deployment option by reducing the Actor
from 27B to 9B parameters while reusing CarryOn's selector and retention
policy. Distillation trains the local Actor to generate actions and
update task state from the same state--evidence interface.

Table~\ref{tab:carryon_e_results} reports 47.0\%, 46.3\%, and 67.5\%
Success on AppWorld, OfficeBench, and Tau2 Telecom, respectively.
On AppWorld, the 9B Actor completes 70.2\% of easy tasks and 62.5\% of
medium tasks, showing that it can handle many tasks at these difficulty
levels. On Tau2, it completes 75.0\% of MMS tasks and 88.9\% of mobile-data
tasks.

\FloatBarrier
\clearpage
\onecolumn
\raggedbottom
\section{Prompt Templates}
\label{sec:prompt_templates}
We provide the Actor instructions, annotation prompts, and selector input
formats in the accompanying code. Braced placeholders below denote runtime
inputs. Benchmark task descriptions, tool documentation, and domain policies
are supplied by the corresponding environment.

\subsection{AppWorld Actor}
The benchmark's task prompt is followed by selected action--observation
messages in chronological order. The following instruction is appended to
the last user message, with the current state substituted for
\texttt{\{working\_memory\}} (\texttt{(empty)} initially).

\begin{promptblock}
**Bounded working-memory protocol**

Your entire response must remain valid Python.
On the first line, write exactly one single-line Python comment in this form:

# MEMORY: done=...; open=...; next=...; answer_mode=none|value; known_apis=...

Replace this memory every turn and keep the text after `# MEMORY:` within 128 tokens. Preserve task progress, the next intended operation, whether the task asks for a returned value, and API endpoint names already inspected. Do not copy passwords, access tokens, long IDs, lists, or tool receipts into memory; exact values remain available in retained raw evidence. After the memory comment, output one small executable Python action.
Do not add prose or Markdown outside Python.

Current working memory:
{working_memory}
\end{promptblock}

\subsection{OfficeBench Actor}
The system message supplies the environment configuration and available apps.
The task workspace replaces the default path at runtime.

\paragraph{System message.}\mbox{}\par

\begin{promptblock}
Today is {date} ({weekday}). The current time is {time}. You are an AI assistant for user {username}.
You can help solve the task step by step.
You can interact with an operation system and use apps to solve the task.
You must follow the instructions and use the given dictionary format to call APIs.
You can only generate one action at a time.
You can find files for your task in `/testbed/data`. If you do not know the filenames, switch to shell and list the directory.
You have following apps installed in the system:
{introductions}
\end{promptblock}

\paragraph{User message layout.}\mbox{}\par

\begin{promptblock}
## Task
{task}

## Selected exact execution evidence
{evidence}

## Current action space
{action_instructions}

{execution_context}

## Output protocol
{output_protocol}
\end{promptblock}

\paragraph{Evidence format.}\mbox{}\par

\begin{promptblock}
[{evidence_id}] ACTION: {action}
[{evidence_id}] OBSERVATION: {observation}
\end{promptblock}

For a partially selected observation, the second line is:

\begin{promptblock}
[{evidence_id}] SELECTED OBSERVATION:
{fragments}
\end{promptblock}

An empty history is rendered as \texttt{(no prior evidence)}.

\paragraph{Action space before selecting an app.}\mbox{}\par

\begin{promptblock}
Available apps: {available_apps}
Choose one app: {'app':'system','action':'switch_app','target_app':APP}
\end{promptblock}

\paragraph{Action space within an app.}\mbox{}\par

\begin{promptblock}
Current app: {current_app}
Available actions:
{action_demos}
- switch app: {'app':'system','action':'switch_app','target_app':APP}; other apps={alternatives}
- finish non-question: {'app':'system','action':'finish_task','answer':'None'}
- finish question: {'app':'system','action':'finish_task','answer':ANSWER}
\end{promptblock}

\paragraph{Execution context.}\mbox{}\par

\begin{promptblock}
## Execution context
Task workspace / default output directory:
"{default_directory}"
Current working directory:
"{current_directory}"
For an output requested by filename only, use the task workspace unless the task specifies another location. Relative filesystem paths resolve from the current working directory.
App-managed storage still follows its tool schema.
Treat file paths, account names and other identifiers as case-sensitive. Copy the exact identity from the task or public evidence; do not substitute the current assistant user for a person named in the task. If unresolved, use an available tool to check rather than guess or normalize spelling.
A successful tool receipt confirms that operation, not all task requirements. Keep uncertain requirements in open. Do not mark the action you are about to execute as already done. Before finish_task, compare the requested outcomes with public receipts: object, location and content. Check missing evidence with existing tools; do not blindly reread every result.
Keep all five MEMORY fields inside one JSON string. Never place done/open/next/answer_mode/known_apis at the outer JSON level.
For shell commands, a string (or singleton string list) is a shell program; a multi-element string list is literal argv.
Redirection and pipes belong in a program string or an explicit shell command, not as literal argv entries.
\end{promptblock}

\paragraph{Output protocol.}\mbox{}\par

\begin{promptblock}
Return exactly one valid JSON object and no prose, Markdown, XML tags, or code fences.
Use this exact outer schema:
{
  "memory":"done=...; open=...; next=...; answer_mode=none|value; known_apis=...",
  "action":{"app":"...","action":"..."}
}
`action` MUST be a nested JSON object, never a quoted string. Include every argument required by one currently available OfficeBench action.
Example switch action:
{
  "memory":"done=none; open=complete the task; next=open shell; answer_mode=none; known_apis=shell",
  "action":{"app":"system","action":"switch_app","target_app":"shell"}
}
The memory must be at most 128 proxy tokens.
Replace it each turn. Keep progress, open work, next operation, answer mode, and recently needed app actions. Never put long IDs, files, lists, or raw receipts in memory; exact values stay in evidence.
Current memory: {working_memory}
\end{promptblock}

The initial memory value is \texttt{none yet}. Without MEMORY, the output instruction is:

\begin{promptblock}
Return exactly one native OfficeBench action dictionary and no prose or Markdown.
\end{promptblock}

\subsection{Tau2 Actor}
The Actor uses Tau2's system instructions and domain policy, followed by the
three CarryOn instruction blocks below. Tool schemas are provided through
the model's native tool interface.

\paragraph{Benchmark system message.}\mbox{}\par

\begin{promptblock}
<instructions>
You are a customer service agent that helps the user according to the <policy> provided below.
In each turn you can either:
- Send a message to the user.
- Make a tool call.
You cannot do both at the same time.

Try to be helpful and always follow the policy. Always make sure you generate valid JSON only.
</instructions>
<policy>
{domain_policy}
</policy>
\end{promptblock}

\paragraph{State update.}\mbox{}\par

\begin{promptblock}
Using working memory without losing the user's goal

MEMORY is a revisable index into the public history, not a substitute for the domain policy or a complete procedure. Keep the existing five fields within 128 tokens, using information available before the current response.

- done: Note completed work and its observed scope. Giving instructions is not the same as the user carrying them out. Attribute user reports as reports.
- open: Preserve the unresolved user outcome, the relevant pending prerequisite or result, and who must act. Finishing your planned checks does not resolve the outcome. Do not drop the goal merely because your previous plan is empty.
- next: State one revisable next step, not a commitment to the previous plan. If an attempted fix leaves the outcome unresolved, reconsider the applicable policy and its prerequisite references. A cause absent from MEMORY has not thereby been ruled out. Before declaring the available procedure exhausted, distinguish a documented limit from a step you have not checked.
- answer_mode and known_apis: Keep their original meanings. Knowing an endpoint does not mean it was called or that its operation succeeded.

Use public results to revise the note. Guide user-owned actions through the documented procedure; do not infer access to private user tools or receipts.
Keep usable confirmations instead of restarting completed checks. When the evidence is sufficient, act directly. Follow the policy when resolution is confirmed or the request needs transfer; do not loop just to keep open nonempty.

Keep the original reply/tool-call protocol. No extra fields, calls or separate verification pass are required.
\end{promptblock}

\paragraph{Historical statements.}\mbox{}\par

\begin{promptblock}
AGENT_STATEMENT records what you previously requested or committed to; it does not certify whether that work is pending or completed. Use the public replies and actual tool receipts to update your existing working memory, retaining the source and scope of confirmations.
\end{promptblock}

\paragraph{Output protocol.}\mbox{}\par

\begin{promptblock}
Bounded CarryOn working-memory protocol:
Every response must provide the same five fields: done, open, next, answer_mode, known_apis. Replace the state each turn. Use 'none' for an empty field.
Keep the complete serialized state within 128 tokens. Preserve task progress, the next intended operation, whether a returned value is needed, and names of tool endpoints already inspected. Do not copy passwords, access tokens, long IDs, lists, or tool receipts into memory. Exact values remain in raw evidence.

There are two output channels for this SAME private control state:
1. Customer reply: start visible text with exactly one line:
# MEMORY: done=...; open=...; next=...; answer_mode=none|value; known_apis=...
Then provide the normal customer-facing reply. A MEMORY line alone is invalid.
2. Native tool call: provide the required _carryon_memory STRING inside the tool arguments, alongside the tool's normal arguments. Its value is:
done=...; open=...; next=...; answer_mode=none|value; known_apis=...
Do not repeat MEMORY in a text block and do not include customer-facing prose on a tool turn. Call exactly one supplied function using its native channel.

The runtime removes private control metadata before giving a reply to the customer or executing the unchanged environment tool. MEMORY is not customer prose, a separate tool action, an observation, or evidence that work succeeded.
Do not output tool calls as Python, JSON text, Markdown, or a code fence.
All existing Tau2 policy and tool-ownership rules apply to the public action.
\end{promptblock}

\paragraph{Tool argument description.}\mbox{}\par

\begin{promptblock}
Required private control state: done=...; open=...; next=...; answer_mode=none|value; known_apis=... . All five fields nonempty; use none when empty. Single line, at most 128 tokens. Removed before tool execution.
\end{promptblock}

Selected evidence is supplied as alternating action and observation messages. The final message also includes:

\begin{promptblock}
Current working memory:
{working_memory}
\end{promptblock}

The initial memory is \texttt{(empty)}. Control metadata is removed before a response reaches the user or a tool executes.

\subsection{Offline State Annotation}
These prompts produce the state annotations and evidence citations used for
selector training. The annotation schema is richer than the Actor's bounded
five-field MEMORY. The following is the shared AppWorld template.

\paragraph{Annotation instruction.}\mbox{}\par

\begin{promptblock}
You label VERIFIED EXECUTION STATE for a tool-using agent from an officially successful AppWorld trajectory.  The trajectory is visible in hindsight, but the state after step t must contain only claims supported by the task and tool evidence at steps <= t.  Return one complete state for every supplied external action, in the same order.  Return JSON only, with no markdown.

Use these terms precisely:
- execution_invariants: long-lived task rules, scope predicates, completeness conditions, preconditions, postconditions, and termination conditions.
- open_obligations: concrete work that remains at this point in the episode.
- verified_facts: environment facts actually supported by tool evidence.
- tool_evidence: concise receipts tying a state claim to one or more step IDs.

Maintain state sequentially rather than independently summarizing each step.
Processed IDs and completed work accumulate unless a later receipt explicitly invalidates them.  Never replace an accumulated set with only the newest item.
Remaining IDs shrink only when there is evidence that an item is processed or out of scope.  Preserve exact entity IDs, record IDs, filenames, pagination cursors, collection members, errors, and mutation receipts when future execution can depend on them.  A successful API response is evidence that the call returned successfully; it is not automatically proof that the overall goal or all side effects are complete.  Mark unverified effects uncertain.

For pagination or enumeration, do not mark exhausted merely because one page was returned.  Exhaustion requires evidence such as an empty terminal page, a documented terminal cursor, or the trajectory's API semantics proving there is no next page.  Keep the current cursor and collected IDs.  For mutations, distinguish intended targets, processed targets, remaining targets, and effects that have been re-queried or otherwise verified.  For answer tasks, preserve the evidence needed to compute the answer and the required answer format.

An execution invariant has status open, satisfied, violated, or unknown.
"satisfied" requires cited evidence; "unknown" is required when evidence is insufficient.  required_before names the action class or condition that cannot safely occur first.  Open obligations are dynamic: add newly discovered work, update it as evidence arrives, and remove or close it only when supported.

state_update describes the transition from the preceding state:
- initialize: the first state;
- update: material state was added, changed, or closed;
- preserve: the latest action adds evidence but does not materially change the maintained state;
- abstain: the observation does not justify a state claim;
- reject: a tempting claim is contradicted by the available evidence.
The added, updated, closed, and rejected arrays contain short field-level descriptions, not hidden reasoning.

termination_supported is true only when no required obligation remains open, all necessary collections are exhausted, required mutations or answers are verified, uncertain effects cannot invalidate success, and the current history contains evidence for the task's completion conditions.  Do not infer it just because the next expert action would complete the task.

The response shape is exactly:
{
  "task_id": "...",
  "initial_state": {
    "step_id": 0,
    "state_update": {
      "mode": "initialize", "added": [], "updated": [],
      "closed": [], "rejected": []
    },
    "execution_invariants": [],
    "verified_facts": {"items": []},
    "tool_evidence": [],
    "collection_state": {
      "processed_ids": [], "remaining_ids": [],
      "next_cursor": null, "exhausted": false
    },
    "open_obligations": [],
    "uncertain_effects": [],
    "termination_supported": false
  },
  "transitions": [
    {
      "step_id": 1,
      "state_update": {
        "mode": "initialize|update|preserve|abstain|reject",
        "added": [], "updated": [], "closed": [], "rejected": []
      },
      "execution_invariants": [
        {
          "invariant": "...",
          "status": "open|satisfied|violated|unknown",
          "evidence_step_ids": [1],
          "required_before": "..."
        }
      ],
      "verified_facts": {
        "items": [
          {
            "fact": "human readable fact name",
            "value": "exact compact JSON or text value",
            "evidence_step_ids": [1]
          }
        ]
      },
      "tool_evidence": [
        {"evidence": "...", "evidence_step_ids": [1]}
      ],
      "collection_state": {
        "processed_ids": [],
        "remaining_ids": [],
        "next_cursor": null,
        "exhausted": false
      },
      "open_obligations": [],
      "uncertain_effects": [],
      "termination_supported": false
    }
  ]
}

Use empty arrays/objects when a category is inapplicable.  Keep states compact but execution-sufficient.  Every evidence_step_ids value must be an integer no larger than the transition's step_id.  Produce exactly one transition for each supplied step and copy its integer step_id exactly.

initial_state is z0=f(task): infer only task-derived execution invariants and open obligations.  Its verified_facts and tool_evidence must be empty, and it must not cite any tool step because no tool action has executed yet.  If an initial invariant uses an evidence_step_ids field for schema consistency, that list must be empty. Represent initial verified_facts exactly as {"items":[]}.
\end{promptblock}

For OfficeBench, \texttt{AppWorld} is replaced by \texttt{OfficeBench}, and the following instruction is appended:

\begin{promptblock}
OfficeBench adaptation: each supplied step is one actually executed native OfficeBench action and its complete public environment observation. Cite these step IDs, not internal lines or list items. Do not invent file contents or effects that the public observations do not establish. Native finish_task is not proof that the task requirements are satisfied.
\end{promptblock}

For Tau2, \texttt{AppWorld} is replaced by \texttt{Tau2 Telecom}, followed by:

\begin{promptblock}
Tau2 public-conversation adaptation: each step contains one actual Actor public response or native tool-call message, followed by all its public tool receipts and public user replies up to the next Actor decision. A step is one indivisible evidence block. Cite whole step IDs, never internal list items. The first public request is the only task instruction supplied. User statements are claims, not verified tool facts. Tool execution does not prove that a user performed an instructed device action. Do not infer hidden device state, private user actions, identity correctness, task success, or evaluator requirements. Use only the public prefix through each labeled step. Future responses must not justify an earlier state. Distinguish unresolved obligations, user reports, and direct tool observations. Terminal dialogue is not verification of success.
\end{promptblock}

\paragraph{Annotation input.}
The user message is a JSON object containing \texttt{task\_id},
\texttt{task\_instruction}, and \texttt{official\_success\_trajectory}.
The trajectory contains \texttt{steps}; Tau2 also includes
\texttt{initial\_public\_prefix}. AppWorld includes
\texttt{tool\_schemas} when available. These fields are filled from the
training trajectory. Output fields and types follow the JSON structure
specified above.

\subsection{Selector Input Format}
The reranker receives a system instruction and a query--document pair.
The query contains the task and current observation; the document contains
the candidate action and observation. The next-action head uses the same
input format with its own task instruction.

\paragraph{Shared system instruction.}\mbox{}\par

\begin{promptblock}
Judge whether the Document meets the requirements based on the Query and the Instruct provided. Note that the answer can only be "yes" or "no".
\end{promptblock}

\paragraph{User input.}\mbox{}\par

\begin{promptblock}
<Instruct>: {instruction}
<Query>: task: {task}
current observation: {current_observation}
<Document>: action: {candidate_action}
observation: {candidate_observation}
\end{promptblock}

OfficeBench reverses the action and observation lines in the document. The input uses Qwen chat delimiters and ends with:

\begin{promptblock}
<|im_end|>
<|im_start|>assistant
<think>

</think>
\end{promptblock}

\paragraph{Ranking: AppWorld and OfficeBench.}\mbox{}\par

\begin{promptblock}
For a long-horizon tool agent, rank whether this exact prior action-observation block should remain visible for correct future execution.
\end{promptblock}

\paragraph{Ranking: Tau2.}\mbox{}\par

\begin{promptblock}
For a long-horizon tool agent, rank whether this exact prior dialogue or tool-evidence unit should remain visible for correct future execution.
\end{promptblock}

\paragraph{Next-action use: AppWorld and OfficeBench.}\mbox{}\par

\begin{promptblock}
For a tool-using agent, judge whether this exact prior evidence supplies an API specification, credential, ID, cursor, path, or runtime value needed by the agent's immediately next executable action.
\end{promptblock}

\paragraph{Next-action use: Tau2.}\mbox{}\par

\begin{promptblock}
For a dual-control dialogue agent, judge whether this exact prior visible evidence is needed by the immediately next decision: to parameterize an agent tool call, issue or complete a user-side instruction or commitment, verify an authoritative effect, or resolve an open obligation. Do not predict general future usefulness.
\end{promptblock}

\subsection{CarryOn-E Teacher Context}
The teacher receives the student's saved context before the action, using
the same Actor instructions and state--evidence interface. Its response is
appended as the supervised assistant target. Initial supervised examples similarly pair the
teacher's recorded context with its response. Training uses the model's
chat template and supervises only the selected assistant responses.

\end{document}